\documentclass[10pt, letterpaper]{article}
\usepackage{authblk}
\usepackage{biblatex} 
\usepackage[a4paper, total={6in, 8in}]{geometry}
\usepackage{float}
\usepackage{hyperref}
\usepackage{longtable}
\usepackage{graphicx}
\usepackage{booktabs}
\usepackage{caption}
\usepackage{subcaption}
\graphicspath{ {./images/} }
\hypersetup{
    colorlinks=true,
    linkcolor=blue,
    filecolor=magenta,      
    urlcolor=cyan,
    pdftitle={Overleaf Example},
    pdfpagemode=FullScreen,
    }
\providecommand{\keywords}[1]
{
  \small	
  \textbf{\textit{Keywords---}} #1
} 

\title{Systematic Literature Review:  Computational Approaches for Humour Style Classification}
\author[1]{Mary Ogbuka Kenneth}
\author[2]{Foaad Khosmood}
\author[3]{Abbas Edalat}

\affil[1,3]{Algorithmic Human Development, Department of Computing, Imperial College London}
\affil[2]{Computer Engineering  Department, California Polytechnic State University}

\begin{document}
\maketitle
\begin{abstract}
Understanding humour styles is crucial for grasping humour's diverse nature and impact on psychology and artificial intelligence. Humour can have both therapeutic and harmful effects, depending on the style. Though computational humour style analysis studies are limited, extensive research exists, particularly in binary humour and sarcasm recognition. This systematic literature review explores computational techniques in these related tasks, revealing their relevance to humour style analysis. It uncovers common approaches, datasets, and metrics, addressing research gaps. The review identifies features and models that can transition smoothly from binary humour and sarcasm recognition to humour style identification. These include incongruity, sentiment analysis, and various models, like neural networks and transformer-based models. Additionally, it provides access to humour-related datasets, aiding future research.
\end{abstract}


\section{Introduction}
Mental health disorders are an increasing public health concern since poor mental health relates to significant emotional distress and contributes significantly to the global illness burden \cite{Bressington2018TheReview}. Duchenne (Genuine) laughter is a common reaction to amusing stimuli or pleasant feelings; it is frequently regarded as having beneficial psychological and physical consequences for humans. As a result, it is considered a possible mental health intervention \cite{Mora-Ripoll2011PotentialResearch}. Laughter therapy, which is a form of cognitive-behavioural therapy, can enhance a person's quality of life overall as well as their physical, psychological, and social connections \cite{Yim2016TherapeuticReview}. The two main types of laughter therapy are humour-based and self-stimulated laughter yoga therapy \cite{Bogodistov2017ProximityDistance, Edalat2021Self-initiatedLaugh}. The use of humour as a therapeutic technique to advance physical, emotional, and psychological well-being is a common theme throughout both therapies.

\par Humour, as a complex and essential aspect of human communication, plays a vital role in social interactions, emotional expression, and cognitive processing \cite{Cann2014SensePerceived,Chandrasekaran2016WeHumor}. Understanding how individuals perceive and express humour and how it relates to our mental health and relationships has garnered significant interest among researchers, leading to the emergence of the field of humour style classification \cite{Saroglou2011BadCouples,Kubert2020IdentifyingStyles,Kuiper2016IdentityWell-Being,Hampes2007TheEmpathy,Kuiper2010PersonalityStyles,Martin2003IndividualQuestionnaire}. Humour style recognition focuses on categorising individuals based on their approaches to humour, shedding light on the diversity and intricacies of humour-related behaviours \cite{Cann2014SensePerceived}. 

Humour style refers to an individual's preferred or characteristic way of expressing humour and engaging in humourous activities. It is a unique and consistent pattern of using humour to interact with others, cope with stress, and express one's personality \cite{Martin2003IndividualQuestionnaire,Veselka2010RelationsPersonality}.  The Humour Styles Questionnaire (HSQ), created by Martin et al. \cite{Martin2003IndividualQuestionnaire}, has identified  four fundamental humour styles, each representing different approaches to humour engagement.   First, there is Affiliative Humour, which is characterised by using humour to foster social bonds and bring people together, with a focus on making others laugh. On the other hand, Self-Enhancing Humour highlights using humour as a coping mechanism, finding amusement in life's challenges, and maintaining a positive outlook. The Aggressive Humour style involves using humour as a form of criticism, sometimes at the expense of others, employing sarcasm or teasing. Lastly, there is Self-Deprecating Humour, where individuals humourously  acknowledge their own imperfections, using self-mockery as a way to connect with others. These four humour styles, as delineated by the HSQ, provide insights into the multifaceted role of humour in social interactions, emotional expression, and coping mechanisms.

 \par The HSQ is a widely used psychological diagnostic tool designed to determine a person's preferred humour style. It typically consists of 32 items, with respondents rating their responses on a Likert scale, indicating their level of agreement or disagreement with each statement \cite{Martin2003IndividualQuestionnaire}.

\par While early studies on humour styles heavily relied on self-report questionnaires, the evolution of computational methods has ushered in automated techniques for humour style identification \cite{Christ2022MultimodalResults,Kamal2020Self-deprecatingApproach,Abulaish2019Self-DeprecatingApproach}. The application of these methodologies in the analysis of extensive textual and multimedia datasets has  enhanced our comprehension of humour styles within various contexts.

\par Although the number of studies specifically focusing on humour style classification using computational models may be limited, there exists a substantial body of research on related tasks, such as binary humour recognition (i.e., determining whether a sentence or scene is humourous or not) and sarcasm recognition. These related tasks share commonalities in humour-related analysis, including feature extraction, representation modelling, and classification algorithms. Understanding the relationship between humour style classification and related tasks is crucial for advancing the field. Lessons learned from humour recognition and sarcasm recognition can be leveraged to improve the accuracy and generalisability of humour style classification models, even with the limited number of specific studies in this domain.

\par There are few review paper \cite{Ramakristanaiah2021ACommunications} and systematic literature review papers that focused on analysis of computational approaches for Humour \cite{Kalloniatis2023ComputationalReview} and sarcasm recognition \cite{Baroiu2022AutomaticReview}. However, there is no review or systematic literature review paper of computational approaches for humour style recognition.

\par This systematic literature review aims to provide a comprehensive analysis of the utilisation of computational approaches for humour style classification and its related tasks. By synthesising the existing literature on humour recognition and sarcasm recognition, the review aims to identify common methodologies, datasets, evaluation metrics, and insights that can inform and enhance humour style classification.

\par The specific objectives of this review are as follows: 
\begin{enumerate}
\item To analyse the computational techniques employed in binary humour recognition and sarcasm recognition.
\item To explore the transferability of  features and computational models from related tasks to humour style classification.\item To identify the strengths and limitations of existing computational-based approaches for humour-related tasks. \item To highlight potential research gaps and opportunities for future investigations in humour style classification using machine learning.
\end{enumerate}

\par By addressing these objectives, this review intends to contribute to the advancement of humour style classification research and its practical applications in various domains, including natural language processing, human-computer interaction, and psychological studies.
\par The rest of this paper is organised as follows: Section 2 describes the review process that was employed to carry out the study. Section 3 displays and explains the outcomes of the review by answering the research questions. Section 4 discusses unresolved challenges and potential future paths in the subject of humour styles and associated tasks. Conclusions were drawn in Section 5.

\section{Methodology}
This section presents the methodology for conducting the SLR on humour style classification and related tasks, such as binary and multi-class humour recognition and sarcasm recognition using computational models. The SLR aims to provide a comprehensive analysis of the existing literature in this domain \cite{Swartz2011TheMeta-analyses}, exploring the computational techniques utilised and the transferability of methodologies from related tasks to humour style classification.

\par The SLR methodology encompasses the following key components: inclusion and exclusion criteria, search strategy and databases used, selection process and study identification, data extraction and analysis, and data synthesis. Through a clear and systematic approach, this methodology enables the review to identify and address the fundamental research questions in this domain\cite{Hordri2017AAnalytics}.

\subsection{Research Questions}
Over the years, a very small number of scholars have used computational methods to accomplish humour style classification. As a result, this study intends to answer the following SLR research questions:
\begin{enumerate}
\item Are there open-source annotated humour styles, sarcasm, and humour datasets in the field of humour recognition?
\item What features are commonly extracted for humour style classification and related tasks, such as humour and sarcasm recognition?
\item What are the computational techniques commonly employed in humour style classification and its related tasks?
\item What are the strengths and limitations of existing computational-based approaches for humour-related tasks, including humour style classification?
\end{enumerate}

\subsection{Search Strategy and Databases}
A systematic and comprehensive literature search was conducted to identify relevant studies. The search strategy involved using appropriate keywords and search terms \cite{Misra2020ADisciplines} related to humour style classification, machine learning, computational models, humour recognition, and sarcasm recognition. \par Examples of the search string used are as follows:
\begin{enumerate}
    \item \textit{(“humour style” OR ``humor style” OR “sarcasm” OR “humour” OR “humor”) AND (“detection” OR “recognition” OR “prediction” OR “classification” OR “identification”)}
    \item \textit{(“machine learning” OR “deep learning” OR ``artificial intelligence") AND (“humour style” OR ``humor style” OR “sarcasm” OR “humour” OR “humor”) AND (“detection” OR “recognition” OR “prediction” OR “classification” OR “identification”)}.
\end{enumerate}

\par The academic databases that were used in the search are: IEEE Xplore (\url{https://ieeexplore.ieee.org/}), ScienceDirect (\url{https://www.sciencedirect.com/}), ACM Digital Library (\url{https://dl.acm.org/}), Scopus (\url{https://www.scopus.com/}), arXiv (\url{https://arxiv.org/}), Semantic Scholar (\url{https://www.semanticscholar.org/} ), Web of Science (\url{https://www.webofscience.com/} ), Google Scholar (\url{https://scholar.google.com/}), Springer Link (\url{https://link.springer.com/} ), and Database systems and logic programming (DBLP) (\url{https://dblp.uni-trier.de/}) 
The search was carried out without any time restrictions to ensure the inclusion of relevant historical studies while focusing on the specified timeframe for inclusion.

\subsection{Inclusion and Exclusion Criteria}
To ensure the relevance and consistency of the selected studies, specific inclusion and exclusion criteria were established. The criteria aimed to capture studies that focused on humour style and related task recognition using computational models, while excluding studies that did not align with the research objectives. The inclusion and exclusion criteria used are depicted in Table \ref{inclusion-exclusion} 

\begin{table}
  \caption{Inclusion and Exclusion Criteria}
  \label{inclusion-exclusion}
  \begin{tabular}{p{0.45\linewidth}| p{0.45\linewidth}}
    \toprule
     \textbf{Inclusion Criteria}& \textbf{Exclusion Criteria}\\
    \midrule
    Studies that addressed humour style classification using computational techniques & Studies that focused solely on non-computational-based humour style classification approaches\\ \hline
    Studies that explored related tasks such as humour recognition and sarcasm recognition & Studies that were not directly related to humour style classification or related tasks \\ \hline
    Studies published in peer-reviewed journals and conference proceedings & Studies published in languages other than English \\ \hline
    Studies written in English & Studies published before the specified timeframe\\ \hline
    Studies published within the last 10 years & Short papers  \\ \hline
    Full-text article &  \\
  \bottomrule
\end{tabular}
\end{table}

\subsection{Selection Process and Study Identification}
The initial search yielded a substantial number of papers related to humour style classification, binary and multi-class humour recognition, and sarcasm recognition. All search results were imported into reference management software to facilitate the removal of duplicates. Subsequently, a title and abstract screening was conducted to identify papers that met the pre-defined inclusion and exclusion criteria.
After the initial screening, the selected papers were obtained in full-text format for further examination.

\subsection{Quality Assessment}
The quality assessment process is used to ensure the rigour and credibility of the primary papers reviewed in this study\cite{Okoli2015AReview}. The quality assessment aimed to evaluate the selected primary studies based on specific criteria to determine their methodological soundness and relevance to the research objectives \cite{Kenneth2022AMAD,Torres-Carrion2018MethodologyEducation}.

\par Quality assessment checklist for the reviewed primary papers:
\begin{enumerate}
    \item \textbf{Research Objectives and Scope:} This checks if the selected studies clearly stated their research objectives and scope in relation to humour style or related task identification. Then we check if the research questions were well-defined and if they were aligned with the overall research objectives of the SLR.
    \item \textbf{Methodology:} Checks if the studies provided a clear description of their approach to humour style or related task recognition. Similarly, the appropriateness of the computational techniques used is checked. Lastly, the suitability of the datasets for the evaluation of the computational techniques is checked. 
    \item \textbf{Evaluation Metrics:} Did the studies employ appropriate evaluation metrics to assess the performance of their machine learning models, and were the evaluation metrics relevant and aligned with the research objectives?
    \item \textbf{Results and Discussion:} Did the studies present their results and findings in a clear and concise manner, and were insights into the effectiveness of the computational models provided? 
    \item \textbf{Limitations and Future Directions:} This checks if the studies acknowledged any limitations in their approach or data and discussed their potential impact on the results. Furthermore, we check if the studies propose future research directions in relation to their work.
    \item \textbf{Citation and References:} Lastly, we check if the studies appropriately cite and reference prior literature relevant to humour and its related tasks.
\end{enumerate}
By conducting a thorough quality assessment, the SLR aimed to present a reliable and comprehensive synthesis of the selected studies and their contributions to the field of humour style classification using computational models and related tasks.

\subsection{Data extraction and analysis}
For the selected studies, a systematic data extraction process was employed to capture relevant information. The data extraction included details such as authors, publication year, study objectives, computational techniques used, datasets employed, evaluation metrics, and key findings. 

\par Following data extraction, a qualitative analysis was performed to identify common methodologies, approaches, and trends across the selected studies. The analysis aimed to provide insights into the computational techniques used in humour and sarcasm classification, highlight their performance, and explore potential transferability to humour style classification.

\par The analysis and conclusions drawn from the selected studies were used to inform potential research gaps and future directions in humour style classification using computational models. The data extraction form that was used to perform an in-depth analysis for all selected primary studies is presented in Table \ref{data-extraction}.

\begin{table}
  \caption{Data Extraction Form}
  \label{data-extraction}
  \begin{tabular}{p{0.25\linewidth}| p{0.45\linewidth}| p{0.20\linewidth}}
    \toprule
    \textbf{Extracted Data} & \textbf{Description} & \textbf{Data Type} \\
    \midrule
     Bibliography & Publication year, author, title and the publisher & General \\ \hline
     Study type & Journal, conference, workshop, or lecture paper & General \\ \hline
     Dataset & Description of dataset used & Research question \\ \hline
     Extracted Features & Description of humorous or sarcastic features extracted and used & Research question \\ \hline
     Computational technique & Employed computational method used like rule-based or machine learning & Research question \\ \hline
     Performance Measure & Performance measures used to evaluate the computational models & Research question \\ \hline
     Findings and contributions & Results obtained and contributions & General \\ 
  \bottomrule
\end{tabular}
\end{table}

\subsection{Data Synthesis}
Data synthesis for the SLR involves analysing and summarising the information extracted from the selected studies on humour style and related task (humour and sarcasm) recognition using computational models.
\par
\begin{figure}[H]
    \begin{subfigure}{0.5\linewidth}
        \centering
        \includegraphics[width=0.8\linewidth]{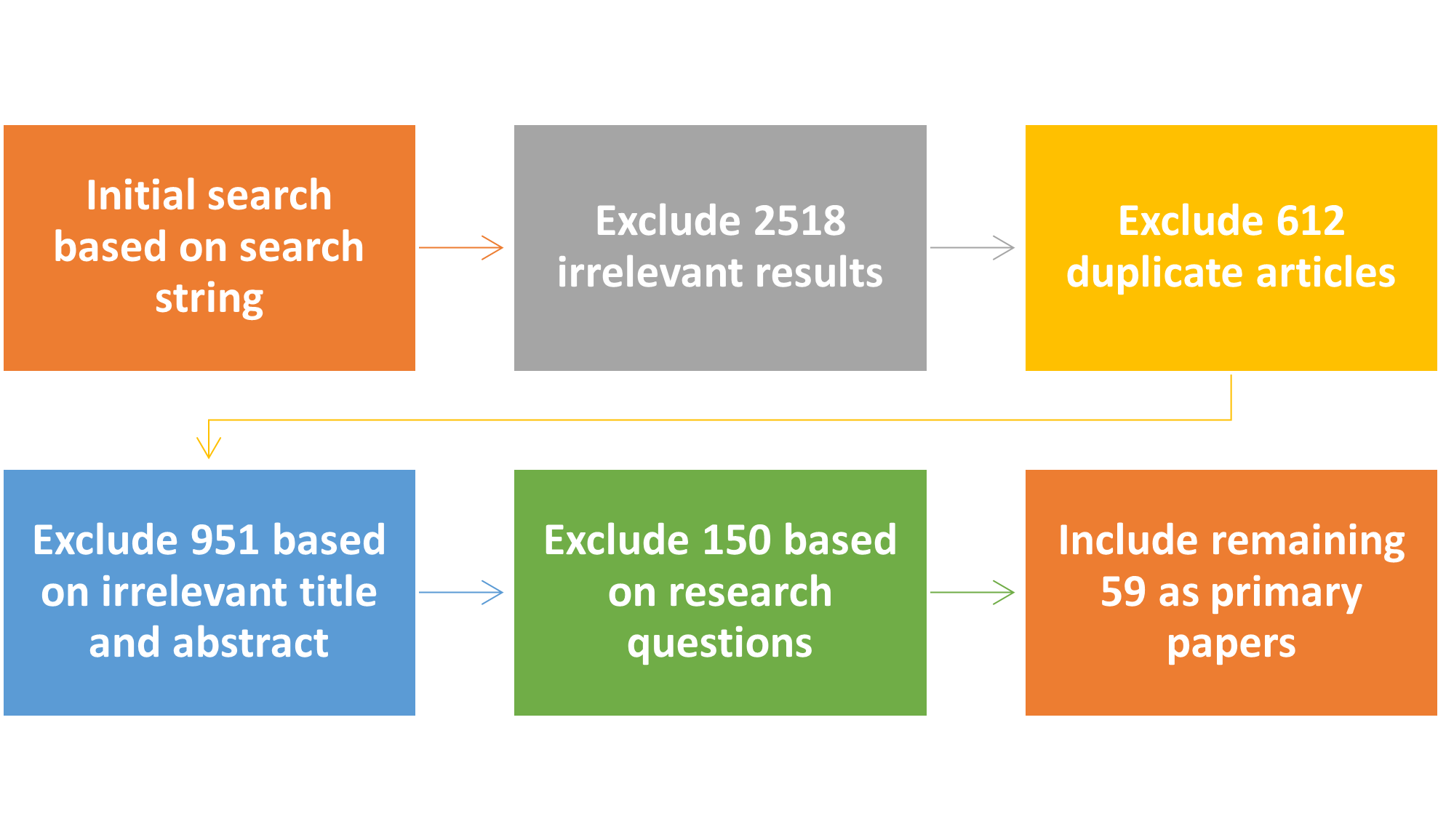}
        \caption{Number of Studies per Systematic Procedure}
        \label{fig:number-of-studies}
    \end{subfigure}%
    \begin{subfigure}{0.5\linewidth}
        \centering
        \includegraphics[width=0.8\linewidth]{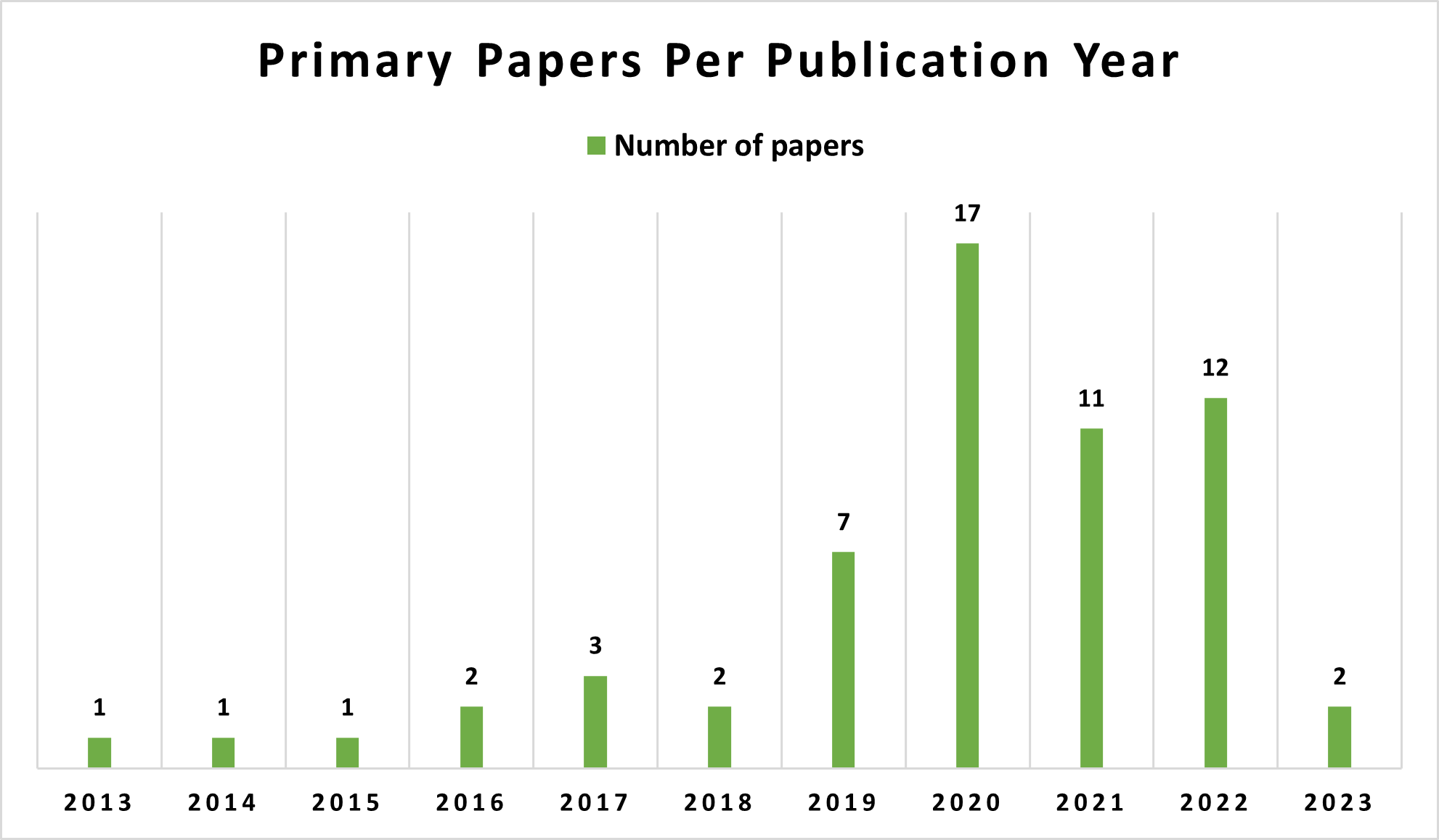}
        \caption{Numbers of papers by year of publication}
        \label{fig:publication-per-year}
    \end{subfigure}
    \caption{Data Synthesis Results}
    \label{fig:data-synthesis}
\end{figure}

 The initial search of the research databases based on the predefined search strings yielded about 4290 studies. Out of these initial results, 2518 were deemed irrelevant, 612 were duplicate articles, 951 were excluded based on irrelevant titles and abstracts, and 209 studies were selected for possible inclusion. Lastly, 150 papers were excluded based on the research question. After following the procedures outlined above, 59 papers were chosen as primary papers. Figure \ref{fig:data-synthesis} shows a visual representation of the results obtained after performing data synthesis.

Figure \ref{fig:number-of-studies} depicts the number of studies conducted for each systematic procedure.
\par The 59 primary papers included were published between 2013 and 2023, spanning a period of ten years. Figure \ref{fig:publication-per-year} displays the number of primary studies by published year. From Figure 

\ref{fig:publication-per-year}, it can be seen that 2020 has the highest selected papers with 17 articles, followed by 2022 with 12 articles, and then 2021 with 11 articles. The years with the fewest primary papers are 2013, 2014, and 2015.

\section{Results and Discussion}
This section presents the results and discussion from the SLR that was conducted to address the stated research questions. Each subsection in this section is dedicated to answering one of the research questions. Additionally, the SLR responses that were gathered from the included primary research paper are reviewed. 

\par A summary of the datasets, extracted features, computational models, and performance metrics used in the 59 primary papers is presented in the appendix in Table \ref{appendix:reviewed-papers}. 
\subsection{Availability of Annotated Datasets}

An essential foundation for advancing research in humour style, sarcasm, and humour recognition lies in the availability of annotated datasets. These datasets serve as crucial resources that enable the development, training, and evaluation of computational models for understanding and identifying humour-related behaviours. 

\par Through a systematic search and selection process, a total of 39 datasets curated for humour style classification, sarcasm recognition, and humour recognition tasks were identified. The datasets are discussed in the context of the annotation techniques used and the dataset modality.

\subsubsection{Dataset Annotation}
The two primary methods utilised for building the datasets in the area of humour styles and their related tasks are manual annotation and the distant supervision technique.
\begin{enumerate}
    \item \textbf{Manual Annotation: }In this method, human annotators review each piece of text (such as jokes, tweets, or comments) and manually assign labels indicating whether it contains humour or sarcasm. This approach allows for detailed analysis and accurate labelling, but it can be time-consuming and subject to annotator bias. Nonetheless, this bias can be reduced by allowing multiple annotators to independently label each instance. The final label is then determined by a majority vote among annotators. This majority vote approach helps mitigate errors and subjectivity. The manual annotation method was greatly adapted for labelling multimodal datasets \cite{Patro2021MultimodalSitcoms,Hossain2019PresidentHeadlines,Castro2018AAnalysis,Christ2022MultimodalResults,KamrulHasan2019UR-FUNNY:Humor,Joshi2016HarnessingFriends}. Riloff et al. \cite{Riloff2013SarcasmSituation} used manual annotation to label the text dataset.

    \item \textbf{Distant Supervision (DS): }In DS, labels are assigned to data instances indirectly based on the presence of some external information or signal. This approach leverages existing knowledge or sources of information to automatically generate labels for a large amount of data. In the context of humour and sarcasm detection, distant supervision involves using emoticons, hashtags, or metadata (such as the context of the text) to infer the label of a text snippet. Most data extracted from Twitter and Reddit made use of this annotation method, as these social media platforms make use of hashtags and emoticons to label text context \cite{Kamal2020Self-deprecatingApproach,Abulaish2019Self-DeprecatingApproach,Weller2020TheCollection,Oprea2019ISarcasm:Sarcasm,Abu-Farha2020FromDataset,Ghosh2016FrackingNetwork}. DS is advantageous in generating very large labelled data for training machine learning models as it is not resource-intensive. However, distant supervision comes with certain challenges, such as limited nuance, noisy labels, or inadequate generalisation.
\end{enumerate}

Out of the 39 identified datasets, 22 were manually labelled, 15  were labelled using distant supervision, and 2 were labelled with a combination of both methods. 

\paragraph{Single Modality Dataset}
\subparagraph{}
\par In the context of humour-related tasks, a single modality dataset typically consists of data from a single mode, such as text, audio, or images, for humour analysis. This means that the dataset captures humour-related information from only one source, limiting the analysis to a specific modality. Examples of single modality datasets are text-only, image-only, or audio-only datasets. 
\par The text modality is the most commonly used single modality for humour and its related tasks. Out of 29 single modality datasets, 28 are text modality \cite{Abulaish2019Self-DeprecatingApproach,Kamal2020Self-deprecatingApproach,Chawla2019ShortDetection,Annamoradnejad2020ColBERT:Humor,Weller2020TheCollection,Ptacek2014SarcasmTwitter,Oliveira2020CorporaPortuguese} and one dataset is image-only modality \cite{Das2018SarcasmCNN}. 
Table \ref{single-modality} gives details of single modality datasets used in the literature.

\begin{longtable}
    {p{0.06\linewidth}|
    p{0.07\linewidth}| 
    p{0.10\linewidth}| p{0.07\linewidth}| p{0.06\linewidth}| p{0.08\linewidth}| p{0.13\linewidth}|
    p{0.06\linewidth} |
    p{0.14\linewidth}}

    \caption{\label{single-modality} Single Modality Datasets} \\ \hline
    \textbf{Paper} & \textbf{Task} 
    & \textbf{Name} & \textbf{Annota-tion}
    & \textbf{Labels} & \textbf{Language} & \textbf{Source} & \textbf{Size} & \textbf{Availability}\\ \hline
    \endfirsthead

    \caption{\label{single-modality} Single Modality Datasets} \\ \hline
    \textbf{Paper} & \textbf{Task} 
    & \textbf{Name} & \textbf{Annota-tion}
    & \textbf{Labels} & \textbf{Language} & \textbf{Source} & \textbf{Size} & \textbf{Availability}\\ \hline
    \endhead

\cite{Kamal2020Self-deprecatingApproach} & Humour & E-tweets & DS & 2 & English & Twitter & 1000 & Tweets IDs are available\\ \hline
    \cite{Abulaish2019Self-DeprecatingApproach} & Sarcasm & E-tweets & DS & 2 & English & Twitter & 1000 & Tweets IDs are available \\ \hline
    \cite{Annamoradnejad2020ColBERT:Humor} & Humour & ColBert & DS & 2 & English & Reddit & 200,000 & \url{https://rb.gy/tif35}\\ \hline
    \cite{Weller2020TheCollection,Huang2022SICKNet:Knowledge}  & Humour & rJokes & DS & 10 & English & Reddit & 550,000 & \url{https://rb.gy/54fcp} \\ \hline
    \cite{Chawla2019ShortDetection} & Humour & Short text & & 6 & English & Different sources & 32,980 & \url{https://rb.gy/22ksm} \\ \hline
    \cite{Ptacek2014SarcasmTwitter,Ren2020SarcasmNetwork} & Sarcasm & Tweets & DS, Manual & 2 & Czech and English & Twitter & 7,000 & \url{https://rb.gy/22ksm} \\ \hline
    \cite{Oliveira2020CorporaPortuguese} & Humour & Portuguese jokes & DS & 2 & Portuguese & Twitter, Newspaper & 2,800 & \url{https://rb.gy/51cxx} \\ \hline
    \cite{Abu-Farha2020FromDataset} & Sarcasm, Sentiment & Arsarcasm & Manual & 6 & Arabic & Twitter & 10,547 & \url{https://rb.gy/r7fev} \\ \hline
     \cite{Oprea2019ISarcasm:Sarcasm} & Sarcasm & iSarcasm & Manual & 8 & English & Twitter & 4,484  & \url{https://rb.gy/ghp2k} \\ \hline
     \cite{Diao2020ADetection} & Sarcasm & IACv2 & Manual & 2 & English & Internet Argument Corpus & 4,692 & \url{https://nlds.soe.ucsc.edu/sarcasm1} \\ \hline
     \cite{Khodak2018ASarcasm,Ahuja2022Transformer-BasedIrony,Kumar2020SarcasmLSTM} & Sarcasm & DS & SARC & 2 & English & Reddit & 175,058 & \url{http://nlp.cs.princeton.edu/SARC/} \\ \hline
     \cite{Misra2023SarcasmDataset, Jariwala2020OptimalHeadlines} & Sarcasm & News headlines & DS & 2 & English & The Onion, HuffPost & 26,709 & \url{https://rishabhmisra.github.io/publications/} \\ \hline
     \cite{Kamal2022CAT-BiGRU:Detection} & Sarcasm & Twitter-280 & DS & 2 & English & Twitter & 42,622 & \url{https://rb.gy/io2xa} \\ \hline
     \cite{Blinov2019LargeRecognition} & Humour & Stierlitz and FUN & DS & 2 & Russian & VK, Twitter, anekdot & 312,877 & \url{https://rb.gy/ctxj8} \\ \hline 
     \cite{Mihalcea2005MakingRecognition,Fan2020HumorNetwork,Fan2020PhoneticsRecognition, Zhang2017InvestigationsRecognition, Ren2021ABML:Detection} & Humour & 1600 One-liner & DS & 2 & English & Berro, Muted Faith, Proverbs, British National Corpus, Reuters & 16,000 & Contact Author \\ \hline
     \cite{Yang2015HumorExtraction,Fan2020HumorNetwork,Fan2020PhoneticsRecognition} & Humour & Pun of the Day & DS & 2 & English & Pun of the Day, News, Proverbs & 4,826 &\url{https://rb.gy/06lgn} \\ \hline
     \cite{Ziser2020HumorSystems} & Humour & Product Question Answering & Manual & 2 & English & Product Question Answering Systems & 28,716 & \url{https://rb.gy/t9bd2/} \\ \hline
     
    \cite{Meaney2021SemEval-2021Offense,Xie2021UncertaintyRecognition} & Humour & SemEval-2021 Task 7 & Manual & 2 & English & Twitter, Kaggle &  10,000 & \url{https://semeval.github.io/SemEval2021/} \\ \hline
     \cite{Pungas2017AJokes,Huang2022SICKNet:Knowledge} & Humour & English Plaintext Jokes & & 2 & English & Reddit, stupidstuff, wocka &  208,000 & \url{https://rb.gy/9y453} \\ \hline
    \cite{Morales2017IdentifyingSources} & Humour & Yelp Dataset & DS & 3 & English & Yelp &  34,000 & \url{https://rb.gy/q2y1y} \\ \hline
    \cite{Hossain2019PresidentHeadlines} : \cite{Cao2021Self-AttentionAssessment,Alexandru2021TracingHeadlines} & Humour & Humicro-edit & Manual & 4 & English & Reddit &  5,000 & \url{https://rb.gy/u0nu3} \\ \hline
    \cite{Castro2018AAnalysis} & Humour & Spanish Corpus & Manual & 2 & Spanish & Twitter &  27,000 & \url{https://t.ly/WmKRF} \\ \hline
    \cite{Swami2018ADetection,Pandey2023BERT-LSTMPost} & Sarcasm & English- Hindi code-mixed & DS, Manual & 2 & English, Hindi & Twitter & 5,250 & \url{https://t.ly/b1acR} \\ \hline
    \cite{Tang2022TheHumor} & Humour & Naughty-former & DS & 4 & English & Reddit, Reuters & 92,153 & \url{https://rb.gy/99tvq} \\ \hline
    \cite{Oraby2017CreatingDialogue,Ren2020SarcasmNetwork} & Sarcasm & IAC & Manual & 2 & English & Internet Argument Corpus & 9,386 & \url{https://rb.gy/99tvq} \\ \hline
    \cite{Ghosh2016FrackingNetwork,Chia2021MachineDetection,Potamias2020ADetection,Eke2021Context-BasedModel} & Sarcasm & Sarcasm-Detection & DS & 2 & English & Twitter & 51,189 & \url{https://rb.gy/ynvzw} \\ \hline
    \cite{Riloff2013SarcasmSituation,Ahuja2022Transformer-BasedIrony,Potamias2020ADetection,Eke2021Context-BasedModel} & Sarcasm & Tweet-Sarcasm & Manual & 2 & English & Twitter & 3,200 & \url{https://rb.gy/jftmz} \\ \hline
    \cite{Das2018SarcasmCNN} & Sarcasm & Yahoo Flickr Sarcasm & DS & 2 & Image & Flickr & 1,846 & \url{https://rb.gy/na3jx}\\ \hline
    
\end{longtable}

\par
Overall, single modality datasets are valuable resources for specific types of humour analysis tasks, especially those that primarily rely on one type of data. Nonetheless, it is essential to acknowledge that humour is a multi-faceted and context-rich phenomenon, and in some cases, incorporating multiple modalities may enhance the accuracy and completeness of the humour analysis.

\paragraph{Multimodal Dataset}
\subparagraph{}
\par A multimodal dataset is a type of dataset that contains multiple modes or types of data, each providing different types of information about a particular subject, event, or phenomenon. The goal of creating multimodal datasets is to enable the study of relationships and patterns that can be discovered by analysing multiple types of data simultaneously. Examples of this modality are text-image, text-audio, and text-audio-video datasets. 
In this SLR, ten multimodal datasets were identified. Most of the multimodal datasets consist of video-text-audio modality and they were manually annotated. 
Table \ref{multimodal-dataset} gives details of the multimodal datasets used in the primary papers. Table \ref{multimodal-dataset} includes the name of the multimodal dataset, the dataset's creator and papers that used it, the task the dataset can be used for, the annotation method used, the number of categories in the data, the language of the data, the sources from which the raw data was extracted, the modality contained in the dataset, and the URL to the dataset, if it is publicly available.

\begin{longtable}
    {p{0.057\linewidth}| p{0.087\linewidth}| 
    p{0.090\linewidth}| p{0.071\linewidth} | p{0.055\linewidth}|
    p{0.078\linewidth}| p{0.174\linewidth}|
    p{0.08\linewidth} |
    p{0.122\linewidth}}
    \caption{\label{multimodal-dataset} Multimodal Datasets}  \\ \hline

    \textbf{Paper} & \textbf{Task} 
    & \textbf{Name} & \textbf{Annota-tion}
    & \textbf{Labels} & \textbf{Lang-uage} & \textbf{Source} & \textbf{Type} & \textbf{Availability}\\ 
    \hline
    \cite{Patro2021MultimodalSitcoms,Liu2022FunnyNet:Videos} & Humour & MHD & Manual & 2 & English & TV show Big Bang Theory & Video, Text & \url{https://tinyurl.com/4a2d3t3x}\\ \hline
    \cite{Castro2019TowardsPaper,Hasan2021HumorHumor,Liu2022FunnyNet:Videos} & Sarcasm & MUStARD & Manual & 2 & English & TV show Friends, Golden Girls, Big Bang Theory, Sarcasmaholics Anonymous. & Video, Text, Audio & \url{https://rb.gy/iwm5p} \\ \hline
    \cite{Chauhan2022AnDetection} & Sarcasm & SEEmoji MUStARD & Manual & 2 & English & TV show Friends, Golden Girls, Big Bang Theory, Sarcasmaholics Anonymous. & Video, Emoji, Text, Audio & \url{https://rb.gy/x8q6l} \\ \hline
    \cite{Christ2022MultimodalResults,Xu2022HybridDetection,Kathan2022AFairness,Christ2022TheStress} & Humour Styles & Passau-SFCH & Manual & 4 & German & Football press conference & Video, Text, Audio & Available on request \\ \hline
    \cite{KamrulHasan2019UR-FUNNY:Humor,Choube2020PunchlineFusion,Hasan2021HumorHumor,Liu2022FunnyNet:Videos} & Humour & UR-FUNNY & Manual & 2 & English & Ted Talks & Video, Text, Audio & \url{https://rb.gy/r6xdp} \\ \hline
    \cite{Li2022Memeplate:Templates} & Humour & Memeplate & Manual & 2 & Chinese & Weibo, Tieba, Baidu, Yandex & Image, Text & \url{https://rb.gy/rnvqe} \\ \hline
    \cite{Sharma2020SemEval-2020Metaphor, Chauhan2020All-in-One:Memes} & Humour, Sarcasm, offensive, motivation  & Memotion & Manual & 4 & English & Various sources & Image, Text & \url{https://rb.gy/2mm2e} \\ \hline
    \cite{Bedi2021Multi-modalConversations} & Humour, sarcasm  & MaSaC & Manual & 4 &  Hindi, English & Tv show (Sarabhai versus Sarabhai) & Audio, Text & \url{https://rb.gy/x27c3} \\ \hline
    \cite{Joshi2016HarnessingFriends,Liu2022FunnyNet:Videos} & sarcasm  & Friends & Manual & 2 &  English & Tv show & Video, Audio, Text & Available on request \\ \hline
    \cite{Chauhan2021M2H2:Conversations} & Humour  & M2H2 & Manual & 2 &  Hindi & Shrimaan Shrimati Phir Se TV show & Video, Audio, Text & \url{https://rb.gy/0jfrv} \\ \hline
\end{longtable}

\subsection{ Commonly Extracted Features}
In the context of humour style classification and related tasks, various features are extracted from the single or multimodal data to capture different aspects of humour or sarcasm. These features can be broadly categorised into three main groups: language-based features, sentiment and tone features, and text representation features. 

\subsubsection{Language-Based Features} These features involve the linguistic elements of humour that contribute to its comedic effect. They play with the structure, context, and style of language to create humorous situations. In this SLR, eleven language-based features such as punctuation, semantic similarity, n-grams and more were identified. 
\begin{enumerate}
    \item \textbf{Contextual Information:} This feature plays a vital role in humour comprehension and interpretation. Humour is often dependent on cultural, social, and linguistic factors, which shape how individuals perceive and respond to humorous content. Key aspects of contextual information include cultural differences, wordplay and language, shared knowledge, social situations, stereotypes, and incongruity. Zhang et al. \cite{Zhang2017InvestigationsRecognition} extracted the latent contextual knowledge at the word level by assigning a contextual similarity score to each instance. Huang et al. \cite{Huang2022SICKNet:Knowledge} extracted contextual information in the form of commonsense features.
    \item \textbf{Incongruity: } Incongruity is one of the primary theories of humour \cite{Scheel2017HumorHealth, Katz1993AHumour}. Incongruity simply refers to a cognitive phenomenon where there is a disconnect or misalignment between what is expected and what actually occurs. Incongruity theory suggests that people find things funny when there is an unexpected twist or discrepancy, challenging their initial assumptions or mental schemas \cite{Katz1993AHumour, Scheel2017HumorHealth, Huang2022SICKNet:Knowledge}. In order to identify incongruity for humour Yang et al. \cite{Yang2015HumorExtraction} measured the semantic disconnection (disconnection and repetition) in a sentence using the Word2Vec model. Ziser et al. \cite{Ziser2020HumorSystems} identified this feature difference between the question and its corresponding product. Other literature that utilises this feature for humour are \cite{Huang2022SICKNet:Knowledge,Morales2017IdentifyingSources,Cao2021Self-AttentionAssessment}.
    \item \textbf{Ambiguity:} Ambiguity in the context of humour refers to the presence of multiple possible interpretations within a joke. This uncertainty in meaning can lead to humorous outcomes as individuals navigate through different interpretations, often resulting in an unexpected or amusing conclusion \cite{Yang2015HumorExtraction, Fan2020PhoneticsRecognition, Fan2020HumorNetwork, Bekinschtein2011WhyAmbiguity}. Yang et al. \cite{Yang2015HumorExtraction} captured the ambiguity of the sentence using WordNet. Fan et al. \cite{Fan2020PhoneticsRecognition}, Morales and Zhai \cite{Morales2017IdentifyingSources}, Hasan et al. \cite{Hasan2021HumorHumor}, Xie et al. \cite{Xie2021UncertaintyRecognition}, Barbieri et al.\cite{Barbieri2014ModellingApproach} and Fan et al. \cite{Fan2020HumorNetwork} also identified and extracted ambiguity in sentences.
    \item \textbf{Phonetic Style: } The term ``phonetic style" in humour refers to using phonetic features to produce comic effects, such as sounds, rhythms, wordplay, and speech patterns \cite{Yang2015HumorExtraction}. Humour is produced by changing the way words are pronounced and how they sound. An essential component of verbal humour that draws on the auditory and phonological features of language is phonetic style. The alliteration and rhyme features were used by Yang \textit{e al}\cite{Yang2015HumorExtraction} to identify the phonetic style. Alliteration simply refers to the repetition of the same initial consonant sound in a sequence of words to create a playful and rhythmic effect \cite{Yang2015HumorExtraction}. Alliteration was also used by \cite{Morales2017IdentifyingSources,Xie2021UncertaintyRecognition} for humour identification.
    \item \textbf{Exaggeration: }Exaggeration is a humour technique that includes dramatically emphasising or overstating something in order to produce a humourous effect \cite{Kamal2020Self-deprecatingApproach}. It frequently entails exaggerating, intensifying, or enhancing an event, description, or ingredient beyond what would be realistic. Kamal and Abulaish \cite{Kamal2020Self-deprecatingApproach} used the exaggeration feature to identify self-deprecating humour style. The identified exaggeration consists of an intensifier in the form of an adverb, adjective, or interjection, along with its frequency count.
    \item \textbf{Part-of-Speech (POS) : }POS features are linguistic elements that provide information about the grammatical category of words in a sentence. These features play a significant role in humour analysis as they help capture the syntactic structure and grammatical patterns that contribute to the comedic effect \cite{Ptacek2014SarcasmTwitter}. This feature was used by Ptacek et al., \cite{Ptacek2014SarcasmTwitter}, Ghosh and Veale \cite{Ghosh2016FrackingNetwork} and Jariwala \cite{Jariwala2020OptimalHeadlines} for sarcasm detection. 
Kamal and Abulaish \cite{Kamal2020Self-deprecatingApproach} used personal pronouns like ``I", ``me", ``my", and ``we" to identify self-deprecating humour style (which involves intentionally highlighting one's own flaws or vulnerabilities in a lighthearted and often relatable way to evoke laughter from others). Kamal and Abulaish \cite{Abulaish2019Self-DeprecatingApproach} also made use of the POS features to identify self-deprecating sarcasm.
    \item \textbf{N-gram-based features: } N-grams can be used to capture linguistic patterns and context in text analysis \cite{Keselj2003N-gram-basedAttribution}. N-grams help identify recurring word sequences, phrases, and linguistic patterns that contribute to the comedic effect of humour.  N-gram (frequency) was used by Barbieri et al.\cite{Barbieri2014ModellingApproach}, Ptacek et al. \cite{Ptacek2014SarcasmTwitter}, Joshi et al. \cite{Joshi2016HarnessingFriends}, Jariwala \cite{Jariwala2020OptimalHeadlines}, Riloff et al.\cite{Riloff2013SarcasmSituation} and Swami et al. \cite{Swami2018ADetection} for sarcasm detection.
    \item \textbf{Punctuation Features: } In the context of humour, this refers to the utilisation and manipulation of punctuation marks within the text to enhance the comedic effect or convey a specific tone. Punctuation marks, such as commas, periods, exclamation points, question marks, and ellipses, can contribute to the overall humorous intent of a text by influencing the reader's perception of timing, emphasis, and tone. Kumar et al. \cite{Kumar2020SarcasmLSTM} identified that sarcasm employs behavioural aspects like low tones, facial gestures, or exaggeration. These low tones, facial gestures, or exaggerations are expressed using punctuation in sentences. Hence, to detect sarcasm Kumar et al. \cite{Kumar2020SarcasmLSTM}, Ptacek et al. \cite{Ptacek2014SarcasmTwitter}, Jariwala \cite{Jariwala2020OptimalHeadlines} and Barbieri et al. \cite{Barbieri2014ModellingApproach} extracted punctuation features. 
    \item \textbf{Term Frequency-Inverse Document Frequency (TF-IDF): }This is a technique used to quantify the importance of words within a collection of documents. The meaning of the word increases proportionally to the number of times in the text the word appears \cite{Qaiser2018TextDocuments}. Words with high TF-IDF scores in a sentence are likely to be indicative of whether it is humorous or not. Pandey and Singh \cite{Pandey2023BERT-LSTMPost} utilised this feature for sarcasm detection. Alexandru and Gîfu \cite{Alexandru2021TracingHeadlines} and Ahuja and Sharma \cite{Ahuja2022Transformer-BasedIrony} applied it to humour detection. 
    \item \textbf{Semantic similarity: } This refers to numerical values or scores that quantify how similar or related two pieces of text, words, or concepts are in terms of their meaning \cite{Chandrasekaran2020EvolutionSurvey}. This feature aims to capture the degree of likeness in meaning between words. This feature was used by Xie et al.\cite{Xie2021UncertaintyRecognition} for humour recognition. 
    \item \textbf{Out-of-vocabulary (OOV) words : } OOV are words that are not typically found in a natural language processing environment's vocabulary. For humour identification in Portuguese, Oliveira et al. \cite{Oliveira2020CorporaPortuguese} utilised this feature to count the number of terms that a Portuguese word2vec model's pre-trained vocabulary did not include.
\end{enumerate}

\subsubsection{Sentiment and Tone}
Sentiment and tone refer to the emotional aspects of language that contribute to the overall mood of a statement. In humour analysis, understanding the sentiment and tone helps determine whether a statement is intended to be funny, sarcastic, or even offensive.
\begin{enumerate}
    \item \textbf{Polarity: } This is the inherent positive, negative, or neutral nature of a sentiment or opinion expressed in a piece of text, speech, or communication. It involves assessing whether the content conveys a positive or negative sentiment towards a particular subject, topic, or situation. The appearance of a positive meaning word in a negative context or vice versa has greatly been used to identify sarcasm \cite{Jariwala2020OptimalHeadlines}. Zhang et al. \cite{Zhang2017InvestigationsRecognition} extracted the affective polarity which consists of emotional polarity, and intensity. Jariwala \cite{Jariwala2020OptimalHeadlines} used polarity as part of the features for sarcasm detection.
    \item \textbf{Subjectivity: } This refers to the fact that the perception and interpretation of what is humorous can vary from person to person based on their individual preferences, experiences, cultural background, and personal context. Humour is inherently subjective, and what one person finds funny, another might not. This subjectivity adds depth and complexity to the study and analysis of humour. In the paper by Zhang et al. \cite{Zhang2017InvestigationsRecognition} subjectivity was computed in the range of 0 to 1, where 0 represents very objective and 1 represents very subjective. Ziser et al. \cite{Ziser2020HumorSystems} identified the subjectivity feature based on the sentiment polarity.
    \item \textbf{Interpersonal Effect: } Interpersonal effects in terms of humour refer to the impact that humour has on social interactions, relationships, and communication between individuals. One important concept of humour is its social or hostility focus, especially regarding its effect on receivers. The interpersonal effect essentially shows that humour is associated with sentiment and subjectivity. The interpersonal effect feature was extracted by Yang \textit{e al}\cite{Yang2015HumorExtraction} based on the words polarity and subjectivity.
    \item \textbf{Sentiment-based features: } Sentiment refers to the emotional tone or attitude expressed in a piece of text \cite{Pravin2015SurveyWatermarking, Alluri2021MultiExtraction, Xu2022AttentionEvaluation}. It encompasses a range of emotions beyond just positive and negative, including joy, anger, surprise, and more \cite{Alluri2021MultiExtraction}. This feature was used by Bouazizi and Otsuki \cite{Bouazizi2016ATwitter}, Kumar et al. \cite{Kumar2020SarcasmLSTM}, Jariwala \cite{Jariwala2020OptimalHeadlines}, Hasan et al. \cite{Hasan2021HumorHumor}, Ghosh and Veale \cite{Ghosh2016FrackingNetwork}, Joshi et al.\cite{Joshi2016HarnessingFriends}, Riloff et al.\cite{Riloff2013SarcasmSituation} and Du et al. \cite{Du2022AnHabits} for sarcasm detection. 
    \item \textbf{Emoticons: } These are textual symbols created by combining characters and symbols to represent facial expressions, objects, or concepts. It is used to convey feelings or ideas. This feature was used by Swami et al. \cite{Swami2018ADetection} for sarcasm detection.
    \item \textbf{Acoustic Features: } This refers to various attributes of sounds or speech patterns. These features are extracted from audio data and can provide valuable insights into the comedic effect and emotional tone of the content. This feature is used in multi-modal humour identification. Acoustic features encompass aspects such as pitch, intonation, rhythm, tempo, and prosody, which contribute to the overall delivery and interpretation of humour. In addition to textual cues, acoustic features add an extra layer of complexity to the understanding of humour. They help convey nuances that may not be fully captured through text alone. The following articles made use of the acoustic features for humour classification:~\cite{Hasan2021HumorHumor, Xu2022HybridDetection, Choube2020PunchlineFusion, KamrulHasan2019UR-FUNNY:Humor, Castro2019TowardsPaper,Bedi2021Multi-modalConversations,Chauhan2021M2H2:Conversations,Liu2022FunnyNet:Videos}.
    \item \textbf{Visual Features: } This refers to the visual cues and elements present in humorous content, particularly in images, videos, or other visual media. These features play a significant role in enhancing the comedic effect and conveying humour to the audience. Visual humour often relies on the juxtaposition of images, unexpected visuals, or exaggerated expressions to create laughter and amusement. Integrating visual cues with text and audio-based humour analysis provides a more comprehensive understanding of humour across different modalities. The literature that utilised the visual features for humour includes: \cite{Hasan2021HumorHumor, Xu2022HybridDetection, Choube2020PunchlineFusion, KamrulHasan2019UR-FUNNY:Humor, Castro2019TowardsPaper,Li2022Memeplate:Templates,Chauhan2020All-in-One:Memes,Chauhan2021M2H2:Conversations,Liu2022FunnyNet:Videos}
\end{enumerate}
\subsubsection{Text Representation}
Text representation involves transforming textual content into numerical forms that can be processed by machine learning algorithms. In humour analysis, text representation methods help capture the underlying patterns and features that contribute to humour.
\begin{enumerate}  
    \item \textbf{Word Embedding: } This entails the representation of words as dense vectors of real numbers in a continuous vector space. Word embedding captures the semantic relationship between words in a sentence. The papers that made use of word embedding for sarcasm detection include \cite{ Misra2023SarcasmDataset, Ghosh2015SarcasticWords,Chauhan2022AnDetection,Bedi2021Multi-modalConversations,Ren2020SarcasmNetwork,Ghosh2016FrackingNetwork,Eke2021Context-BasedModel, Sharma2020SemEval-2020Metaphor} while the papers that used this feature for humour identification are \cite{Kamal2020Self-deprecatingApproach, Fan2020HumorNetwork, Choube2020PunchlineFusion,Cao2021Self-AttentionAssessment,Chauhan2021M2H2:Conversations,Ahuja2022Transformer-BasedIrony,KamrulHasan2019UR-FUNNY:Humor,Sharma2020SemEval-2020Metaphor}. 
    \item \textbf{Deep Features: } Deep features refer to complex and abstract representations of textual, visual, or auditory content that have been learned by deep learning models. These features capture high-level patterns \cite{Kenneth2022AMAD}, relationships, and representations that are not directly interpretable by humans but are useful for humour and sarcasm identification tasks. For sarcasm detection, the papers that made use of deep features are \cite{Kumar2020SarcasmLSTM, Abu-Farha2020FromDataset, Das2018SarcasmCNN,Ghosh2016FrackingNetwork,Chia2021MachineDetection,Potamias2020ADetection,Sharma2020SemEval-2020Metaphor}. While for humour detection, the literature that used  deep features are \cite{Xu2022HybridDetection, KamrulHasan2019UR-FUNNY:Humor, Castro2019TowardsPaper,Li2022Memeplate:Templates,Chauhan2020All-in-One:Memes,Tang2022TheHumor,Liu2022FunnyNet:Videos,Ahuja2022Transformer-BasedIrony,Sharma2020SemEval-2020Metaphor}
    \item \textbf{Bag of Word (BOW): } is a text representation technique used to represent text data as numerical vectors \cite{Qadar2019AnChallenges}. In BOW, each sentence or phrase is represented as a vector, where each dimension corresponds to a word in the vocabulary, and the value in the vector represents the frequency of that word in the document \cite{Qadar2019AnChallenges}. This feature was used by Ghosh and Veale \cite{Ghosh2016FrackingNetwork} for sarcasm detection. 
\end{enumerate}
Table \ref{feature-extracted} provides a summary of the features used in the primary literature to detect either humour styles, humour, sarcasm, or related-tasks. Table \ref{feature-extracted} includes the major categories of the extracted features, including the features themselves, the tasks they were used for, the papers they were used in, and the overall number of times they were used.

\begin{longtable}
  {p{0.11\linewidth}| p{0.21\linewidth}| p{0.17\linewidth}| p{0.19\linewidth}| p{0.07\linewidth}| p{0.09\linewidth}} 

\caption{\label{feature-extracted} Summary of Extracted Features} 
\\ \hline

    \textbf{Categories} & \textbf{Features} 
    & \textbf{Humour} & \textbf{Sarcasm} & \textbf{Humour Style}
    & \textbf{Number of studies} \\ \hline

    Language-based &Contextual Information& \cite{Zhang2017InvestigationsRecognition} & & & 1 \\ \cline{2-6}
     & Incongruity & \cite{Cao2021Self-AttentionAssessment,Huang2022SICKNet:Knowledge,Morales2017IdentifyingSources,Yang2015HumorExtraction,Ziser2020HumorSystems} &  & & 5  \\ \cline{2-6}
     & Ambiguity & \cite{Fan2020HumorNetwork,Fan2020PhoneticsRecognition,Hasan2021HumorHumor,Morales2017IdentifyingSources,Xie2021UncertaintyRecognition,Yang2015HumorExtraction} & \cite{Barbieri2014ModellingApproach} &  & 7 \\ \cline{2-6}
     & Phonetic Style & \cite{Morales2017IdentifyingSources,Xie2021UncertaintyRecognition,Yang2015HumorExtraction} &  &  & 3\\ \cline{2-6}
     & Exaggeration &  &  & \cite{Kamal2020Self-deprecatingApproach} & 1  \\ \cline{2-6} 
     & Part-of-Speech &  & \cite{Ptacek2014SarcasmTwitter,Jariwala2020OptimalHeadlines,Ghosh2016FrackingNetwork} &\cite{Kamal2020Self-deprecatingApproach,Abulaish2019Self-DeprecatingApproach} & 5\\ \cline{2-6}
     & N-gram &  & \cite{Barbieri2014ModellingApproach,Ptacek2014SarcasmTwitter,Swami2018ADetection,Jariwala2020OptimalHeadlines,Joshi2016HarnessingFriends,Riloff2013SarcasmSituation} &  & 6 \\ \cline{2-6}
     & Punctuation Features &  & \cite{Barbieri2014ModellingApproach,Kumar2020SarcasmLSTM,Ptacek2014SarcasmTwitter,Jariwala2020OptimalHeadlines} & & 4\\ \cline{2-6}
     & TF-IDF & \cite{Alexandru2021TracingHeadlines} & \cite{Pandey2023BERT-LSTMPost} &  & 2 \\ \cline{2-6}
    & Semantic Similarity & \cite{Xie2021UncertaintyRecognition} &  &  & 1 \\ \cline{2-6}
    & OOV words & \cite{Oliveira2020CorporaPortuguese} &  &  & 1 \\  \hline 
    
    Sentiment and Tone & Polarity & \cite{Zhang2017InvestigationsRecognition} & \cite{Jariwala2020OptimalHeadlines}  &  & 2 \\ \cline{2-6}
     & Subjectivity & \cite{Zhang2017InvestigationsRecognition,Ziser2020HumorSystems} & & & 2\\ \cline{2-6}
     & Interpersonal Effect & \cite{Yang2015HumorExtraction} & & & 1 \\ \cline{2-6}
     & Sentiment-based & \cite{Hasan2021HumorHumor}  & \cite{Bouazizi2016ATwitter,Kumar2020SarcasmLSTM,Du2022AnHabits,Jariwala2020OptimalHeadlines,Joshi2016HarnessingFriends,Ghosh2016FrackingNetwork,Riloff2013SarcasmSituation} & & 8\\ \cline{2-6}
     
     & Emoticons &   & \cite{Swami2018ADetection} & & 1\\ \cline{2-6}
     & Acoustic Features & \cite{Choube2020PunchlineFusion,Hasan2021HumorHumor,KamrulHasan2019UR-FUNNY:Humor,Xu2022HybridDetection,Christ2022TheStress,Kathan2022AFairness,Bedi2021Multi-modalConversations,Chauhan2021M2H2:Conversations,Liu2022FunnyNet:Videos}  & \cite{Castro2019TowardsPaper} & \cite{Christ2022MultimodalResults} & 11\\ \cline{2-6}
     & Visual Features & \cite{Choube2020PunchlineFusion,Hasan2021HumorHumor,KamrulHasan2019UR-FUNNY:Humor,Xu2022HybridDetection,ChauhanAll-in-One:Memes,Christ2022TheStress,Kathan2022AFairness,Patro2021MultimodalSitcoms,Chauhan2021M2H2:Conversations,Liu2022FunnyNet:Videos}  & \cite{Castro2019TowardsPaper,ChauhanAll-in-One:Memes,Kumar2020SarcasmLSTM} & \cite{Christ2022MultimodalResults} & 14  \\ \hline 
     
    Text Representation & Word Embedding & \cite{Cao2021Self-AttentionAssessment,Choube2020PunchlineFusion,Fan2020HumorNetwork,Kamal2020Self-deprecatingApproach,Bedi2021Multi-modalConversations,Chauhan2021M2H2:Conversations,KamrulHasan2019UR-FUNNY:Humor,Choube2020PunchlineFusion, Sharma2020SemEval-2020Metaphor} &\cite{Misra2023SarcasmDataset,Ghosh2015SarcasticWords,Ren2020SarcasmNetwork,Eke2021Context-BasedModel, Sharma2020SemEval-2020Metaphor} &  & 14 \\ \cline{2-6}
     & Deep Features & \cite{Castro2019TowardsPaper,Chauhan2020All-in-One:Memes,KamrulHasan2019UR-FUNNY:Humor,Li2022Memeplate:Templates,Xu2022HybridDetection,Christ2022TheStress,Kathan2022AFairness,Patro2021MultimodalSitcoms,Tang2022TheHumor,Liu2022FunnyNet:Videos,Sharma2020SemEval-2020Metaphor} & \cite{Abu-Farha2020FromDataset,Das2018SarcasmCNN,Kumar2020SarcasmLSTM,Ghosh2016FrackingNetwork,Chia2021MachineDetection,Potamias2020ADetection,Sharma2020SemEval-2020Metaphor}& \cite{Christ2022MultimodalResults} & 19  \\ \cline{2-6}
      & Bag of Words &  & \cite{Ghosh2016FrackingNetwork} &  & 1  \\ \hline
\end{longtable}

\subsection{Commonly Employed Computational Techniques}

In this section, we delve into the rich landscape of computational techniques that have been explored in existing works within the realm of humour style classification and related tasks such as humour and sarcasm recognition. These techniques encompass a diverse range of methodologies, from traditional machine learning (ML) approaches to state-of-the-art neural network models, transformer models, and specialised models tailored specifically to humour and sarcasm tasks.

\subsubsection{Traditional Machine Learning (TML) Models} These techniques encompass classical machine learning algorithms. These traditional ML models are well established and interpretable. However, they require manual feature engineering to capture humour-related patterns effectively. The different traditional ML models used in the reviewed papers are Support Vector Machine (SVM), K-nearest neighbour (KNN), Random Forest (RF), Decision Tree (DT), Naive Bayes (NB), Logistic Regression (LR) and ensemble models like Bagging and Boosting. The most commonly used TML method for humour detection is the SVM, with 6 papers using it for binary humour classification, followed by NB with 5 papers, RF with 3 papers, Ensemble with 2 papers, and LR, DT, and KNN with 1 paper each. Also for sarcasm identification, the SVM model was mostly adapted with 11 papers utilising it for sarcasm detection, followed by NB with 3 papers. The KNN, RF, and LR were adapted equally in two papers, each using them for sarcasm identification. The least used TML model was the DT and ensemble, with only 1 paper utilising them. 
\par Furthermore, it was noticed that only three papers \cite{Christ2022MultimodalResults,Joshi2016HarnessingFriends,Castro2019TowardsPaper} utilised the TML models precisely SVM for multi-modal humour/sarcasm identification. For multimodal datasets, the best F1-score obtained by the TML (SVM) model was 79.8\% \cite{Joshi2016HarnessingFriends}. 
\par In the identification of a single humour style (self-deprecating), the DT, RF, and Bagging showed to be effective with F1-scores of 94.8\% \cite{Abulaish2019Self-DeprecatingApproach}, 87\% \cite{Kamal2020Self-deprecatingApproach}, and 93\% \cite{Abulaish2019Self-DeprecatingApproach}, respectively. It can be inferred that all these models have DT in common, either as a single model or as a base model. This shows that DT is well-suited for the detection of self-deprecating and self-enhancing humour styles. Similarly, Christ et al. \cite{Christ2022MultimodalResults} used SVM to identify the direction and sentiment of a joke, given that jokes are assigned humour style labels based on their direction and sentiment. For example, the self-enhancing humour style has a positive sentiment and is self-directed, while the aggressive humour style has a negative sentiment and is others-directed. 
\par Overall, the SVM performed well across several papers, with an F1-score range of 93\% \cite{Ptacek2014SarcasmTwitter} to 51\% \cite{Riloff2013SarcasmSituation} for binary sarcasm identification and up to 88\% \cite{Annamoradnejad2020ColBERT:Humor} F1-score for binary humour. The least performing model was KNN with an F1-score of 31\% \cite{Ahuja2022Transformer-BasedIrony} for sarcasm detection. 
\par In conclusion, a total of 19 reviewed papers made use of TML models for humour detection, while 22 reviewed papers used them for sarcasm analysis. In addition, the SVM model was the most commonly used model for both sarcasm and humour identification in text datasets because it is effective in high-dimensional spaces, which makes it well-suited for text classification, and it is also robust to overfitting as it can generalise well to unseen data when trained with a small dataset.

\subsubsection{Neural Network (NN) Models} NN models have gained prominence for their ability to automatically learn complex patterns from data. These models excel at capturing sequential and contextual information, making them suitable for humour or sarcasm analysis tasks. The NN models adapted in the 59 reviewed papers are: Long short-term memory (LSTM), Convolutional Neural Network (CNN), Gated Recurrent Unit (GRU), Recurrent Neural Networks (RNN), Artificial Neural Network (ANN) and Bidirectional-LSTM. 
The most used NN model for humour task was LSTM, with 8 reviewed papers using it, followed by CNN with 4 reviewed papers, whereas CNN was the most used for the sarcasm task, with 7 reviewed papers utilising it as compared to LSTM with only 4 papers using it for the sarcasm task.
\par Out of the 59 reviewed papers, 17  made use of multimodal datasets, and out of these 17 papers, 11 made use of neural network (NN) models for humour-related task. This shows that the NN is the most commonly used computational model for multimodal humour or sarcasm detection. Looking at the NN performance for multimodal datasets, the LSTM model had the best performance with an F1-score of 81.4\% \cite{Bedi2021Multi-modalConversations} for humour and 68.6\% \cite{Bedi2021Multi-modalConversations} for sarcasm detection. The next best performing model is the RNN model, with an F1-score of up to 76.7\% \cite{Chauhan2022AnDetection}.  
\par Considering the single modality models, 11 papers made use of NN modals for sarcasm or humour detection on text, while one paper used it on an image dataset \cite{Das2018SarcasmCNN}. Kamal and Abulaish \cite{Kamal2019AnData} used LSTM for detection of self-deprecating sarcasm in text. Whereas, Das and Anthony \cite{Das2018SarcasmCNN} used the CNN model to identify sarcasm in images. The CNN model obtained an accuracy of 84\% for this task \cite{Das2018SarcasmCNN}. The LSTM and CNN models showed promising performance for text modality, with both achieving an accuracy of 84\% \cite{Ziser2020HumorSystems} for the humour classification task and an F1-score and accuracy ranging from 81\% to 96\% sarcasm-related task \cite{Pandey2023BERT-LSTMPost,Ahuja2022Transformer-BasedIrony}.

\subsubsection{Transformer-Based (TB) Models}
These models are built upon a unique architecture that employs self-attention mechanisms and feed-forward neural networks, allowing them to capture complex dependencies and relationships in data. The TB models used in the included literature are as follows: Bidirectional Encoder Representations from Transformers (BERT) \cite{Patro2021MultimodalSitcoms,Pandey2023BERT-LSTMPost,Ahuja2022Transformer-BasedIrony,Tang2022TheHumor}, Robustly Optimised BERT (RoBERTa) \cite{Li2022Memeplate:Templates,Ahuja2022Transformer-BasedIrony,Tang2022TheHumor}, Decoding-enhanced BERT with disentangled attention (DeBERTa) \cite{Tang2022TheHumor}, XLNet \cite{Hasan2021HumorHumor,Ahuja2022Transformer-BasedIrony}, GPT-2 \cite{Xie2021UncertaintyRecognition}and A Lite BERT (ALBERT) \cite{Hasan2021HumorHumor}. 
Looking at the multimodal datasets, the transformer-based models (TBM) were adapted by only 3 papers \cite{Patro2021MultimodalSitcoms,Li2022Memeplate:Templates,Hasan2021HumorHumor} and these papers mostly used them as the baseline models for comparison and not as the main classification models. Nonetheless, the TB models produced good classification results with BERT obtaining an accuracy of 67.98\% \cite{Patro2021MultimodalSitcoms} and 69.62\% \cite{Hasan2021HumorHumor} and XLNet having an accuracy of up to 72.43\% \cite{Hasan2021HumorHumor}. Li et al. \cite{Li2022Memeplate:Templates} used a combination of two transformer models for classification. The combinations include RoBERTa + cross-covariance image transformer (XCiT) and RoBERTa + bidirectional encoder representation from image transformers (BEiT). 
\par Out of the 40 included papers that performed humour or sarcasm detection on text datasets 9 papers utilised only a single TB model without a combination with other model types \cite{Ahuja2022Transformer-BasedIrony,Tang2022TheHumor,Pandey2023BERT-LSTMPost,Cao2021Self-AttentionAssessment,Annamoradnejad2020ColBERT:Humor,Huang2022SICKNet:Knowledge,Weller2020TheCollection,Xie2021UncertaintyRecognition,Meaney2021SemEval-2021Offense}. Two papers \cite{Cao2021Self-AttentionAssessment,Weller2020TheCollection} used TB models BERT and RoBERTa as Regressors to predict the level of funniness of a humour. Weller and Kelvin \cite{Weller2020TheCollection} predicted the level of funniness of a joke on a scale of 0 to 10 and then got a Root Mean Square Error (RMSE) of 1.619, 1.614 and 1.739 for BERT, RoBERTa, and XLNet, respectively.  On the other hand, Cao \cite{Cao2021Self-AttentionAssessment} predicted the funniness of an edited news headline on a scale of 0 to 3 and got RMSE scores of 0.528 and 0.516 for BERT and RoBERTa, respectively. The TB models performed significantly well across the remaining 7 papers that used them as classifiers. For instance, in the work by Annamoradnejad and Zoghi \cite{Annamoradnejad2020ColBERT:Humor} BERT got an accuracy and F1-score as high as 98.2\% and XLNet got an accuracy of 91.6\% and F1-score of 92\%. Meaney et al. \cite{Meaney2021SemEval-2021Offense} BERT's model obtained a 91\% accuracy and 92.8\% f1-score, while Pandey and Singh \cite{Pandey2023BERT-LSTMPost} got an F1-score of 97\% using BERT. Given the recent popularity of GPT models, only Xie et al. \cite{Xie2021UncertaintyRecognition} used GPT-2 for humour detection. Their GPT-2 model got an accuracy of 83.7\% and an F1-score of 83.6\%.
\par Overall, given the performance of TB models on humour-related tasks, it can be concluded that they are well-suited for text-based humour or sarcasm recognition tasks.

Table \ref{tab:computational-models} summarises the computational models that were applied in the initial research on humour and its related task recognition. The table lists the computational models that were employed, the tasks for which they were utilised, and the number of studies that did so. The \# in the table means number.

\begin{table}[h!]
\caption{Summary of Studies and Computational Model Types}
\label{tab:computational-models}
\begin{tabular}
{p{0.16\linewidth}| p{0.21\linewidth}| p{0.21\linewidth}| 
p{0.25\linewidth}}

\hline
\multicolumn{4}{|c|}{\textbf{Traditional Machine Learning Models}} \\ \hline
\textbf{Model} & \textbf{Sarcasm \# of studies} & \textbf{Humour \# of studies} & \textbf{Humour Styles \# of studies}\\
\midrule
 SVM & 11 & 6 & 1 \\ 
NB  & 3 & 5 & 1 \\ 
RF  & 2 & 3 & 1 \\ 
LR  & 2 & 1 & - \\ 
DT  & 1 & 1 & 1 \\ 
KNN & 2 & 1 & - \\ 
Others &1& 2& 1 \\ \hline
\toprule
\multicolumn{4}{|c|}{\textbf{Neural Networks Models}} \\ \hline
\textbf{Model} & \textbf{Sarcasm \# of studies} & \textbf{Humour \# of studies} & \textbf{Humour Styles \# of studies}\\
\midrule
LSTM & 5 & 8 & 1 \\ 
CNN  & 7 & 4 & - \\ 
GRU  & 2 & 1 & - \\ 
RNN  & 1 & 2 & - \\ 
ANN  & 1 & 2 & - \\ 
Others& 5 & 2 & -\\

\toprule
\multicolumn{4}{|c|}{\textbf{Transformer-Based Models}} \\ \hline
\textbf{Model} & \textbf{Sarcasm \# of studies} & \textbf{Humour \# of studies} & \textbf{Humour Styles \# of studies}\\
\midrule
BERT     & 2 & 8 & - \\ 
RoBERTa  & 1 & 3 & - \\ 
XLNet    & 1 & 3 & - \\ 
GPT-2    & - & 1 & - \\ 
Others   & - & 2 & - \\ 

\toprule
\multicolumn{4}{|c|}{\textbf{Specialised Models}} \\ \hline
\textbf{Model} & \textbf{Sarcasm \# of studies} & \textbf{Humour \# of studies} & \textbf{Humour Styles \# of studies}\\ 
\midrule
Specialised models &6& 9& - \\
\bottomrule
\end{tabular}
\end{table}

\subsubsection{Specialised Models}Some research efforts on humour-related tasks have focused on developing specialised models tailored specifically to humour or sarcasm tasks. These models go beyond general-purpose natural language processing techniques and instead incorporate domain-specific knowledge and linguistic features that are particularly relevant to humour analysis. These specialised models include: Multi-level Memory Network-based on Sentiment Semantics (MMNSS) \cite{Ren2020SarcasmNetwork}, Multimodel Self Attention Model (MSAM) \cite{Patro2021MultimodalSitcoms}, Humour Knowledge Enriched Transformer (HKT) \cite{Hasan2021HumorHumor},Muti-dimension attention-based BiLSTM (MDA-BiLSTM) \cite{Diao2020ADetection}, Context-Aware Hierarchical Network (CAHN) \cite{Choube2020PunchlineFusion}, Modality-Invariant and Specific Representations for Multimodal Sentiment Analysis (MISRMSA) \cite{Chauhan2021M2H2:Conversations}, BERT-LSTM \cite{Pandey2023BERT-LSTMPost}, Semantic Incongruity and Commonsense Knowledge Network (SICKNet) \cite{Huang2022SICKNet:Knowledge}, Multi-dimension Intra-Attention Network (MIARN) \cite{Oprea2019ISarcasm:Sarcasm}, Multi-Head Attention (MHA) \cite{Ren2021ABML:Detection,Cao2021Self-AttentionAssessment}, Contextual Extension of memory fusion Network (C-MFN) \cite{KamrulHasan2019UR-FUNNY:Humor} and Phonetics and Ambiguity Comprehension Gated Attention Network (PACGA) \cite{Fan2020PhoneticsRecognition}. 
\par These specialised models have widely been used in both multimodal and single modality humour or sarcasm detection. The collective findings from 13 included studies employing these specialised models consistently underline their superiority over general-purpose natural language processing techniques. For instance, SICKNet, with its focus on semantic incongruity detection and commonsense knowledge extraction, demonstrated a significant 3\% F1-score improvement over BERT and a remarkable 6\% improvement over RoBERTa on the same humour task, as reported by Huang et al. \cite{Huang2022SICKNet:Knowledge}. Similarly, Hasan et al. \cite{Hasan2021HumorHumor}'s HKT model recorded impressive performance increments of 5\% and 8\% when compared to XLNet and BERT, respectively, for humour detection in multimodal datasets.
\par These findings collectively underscore the importance of specialised models in advancing the state of the art in humour and sarcasm detection. The field benefits from these models' ability to capture nuanced linguistic patterns, leverage domain-specific knowledge, and effectively address the unique challenges posed by humour analysis.

\subsection{Strengths and Limitations of Existing Computational Approaches}
\par Evaluating the strengths and limitations of existing computational-based approaches, encompassing both features and models, within the domain of humour style classification and related tasks yields the following insights:

\subsubsection{Strengths}
\begin{enumerate}
    \item \textbf{Feature Diversity:} The extensive array of features, including contextual information, incongruity, and sentiment-based metrics, offers flexibility in capturing the multifaceted aspects of humour styles. This feature diversity allows for a nuanced analysis of humour expressions.
    \item \textbf{Interpretability:} Traditional ML models such as SVM, NB, and DT provide transparent decision-making processes, aiding in understanding the role of individual features in classifying humour styles. This interpretability can be valuable for model validation and debugging.
    \item \textbf{Contextual Awareness:} Based on the reviewed papers, the transformer-based models, notably BERT, RoBERTa, and GPT-2, excel at grasping contextual nuances, making them well-suited for humour styles that heavily rely on context.
    \item \textbf{Multimodal Capability:} Specialised models like MSAM and MISRMSA cater to multimodal sentiment analysis, enabling the incorporation of acoustic and visual cues into humour style classification. This is particularly beneficial for humour expressed across various modalities.
    \item \textbf{Transferability:} The adaptability of existing models from humour and sarcasm detection to humour style classification demonstrates their robustness and potential for re-purposing. This can accelerate research efforts by building upon established architectures.
\end{enumerate}
\subsubsection{Limitations and Challenges}
    \begin{enumerate}
        \item \textbf{Subjectivity and Ambiguity: } Humour, by its nature, is subjective and context-dependent. Existing models may struggle to capture the diverse interpretations of humour and sarcasm, leading to potential misclassifications or limited generalisations.
        \item \textbf{Data Biases: } Computational models are highly dependent on training data. Biases present in the training data can perpetuate stereotypes or under-represent certain humour styles, sarcasm, affecting model fairness and performance.
        \item \textbf{Generalisation:} Models may perform well on specific humour types seen in the training data but struggle with humour styles that exhibit unique linguistic or contextual characteristics. Achieving robust generalisation remains a challenge.
        \item \textbf{Multimodal Integration:} While specialised models cater to multimodal data, integrating features from different modalities seamlessly remains a complex task. Ensuring a balanced contribution from textual, acoustic, and visual features is challenging.
        \item \textbf{Model Overhead:} Some advanced models, particularly transformer-based models, may require substantial computational resources and longer training times, limiting their accessibility for researchers with limited computational capabilities.
    \end{enumerate}
\subsubsection{Areas for Improvement}
    \begin{enumerate}
        \item \textbf{Diverse Datasets:} Developing diverse and balanced datasets that encompass various humour styles is essential for training models that generalise well across a wide range of comedic expressions.
        \item \textbf{Bias Mitigation:} Implementing techniques to mitigate biases in training data and models is critical for ensuring fairness in humour style and related-task classification.
        \item \textbf{Interpretable Models:} Enhancing the interpretability of complex models like transformers can aid in understanding their decision-making processes and improving model trustworthiness.  
        \item \textbf{Multimodal Fusion:} Developing effective strategies for integrating textual, acoustic, and visual features to create a holistic understanding of humour styles is an area ripe for innovation.
        \item \textbf{Transfer Learning:} Investigating transfer learning approaches to leverage pre-trained models for humour style and related-task classification can optimise resource utilisation and enhance performance.
    \end{enumerate}
In conclusion, the strengths and limitations of existing computational-based approaches in humour style and related-task classification highlight the multifaceted nature of this task. Addressing these challenges while capitalising on their strengths will advance the state of the art in computational humour analysis and its applications across various domains.

\section{Transferable Features and Models for Humour Style Recognition}
Drawing from the established features and computational models within the domains of humour styles, binary humour recognition, and sarcasm detection, several features and models can be inferred to be potentially transferable for humour style recognition. These identified features and models provide a solid groundwork upon which to construct and assess computational methodologies tailored for the specific challenge of humour style classification.
\subsection{Transferable Features} 
\begin{enumerate}
        \item \textbf{Contextual Information:} Context plays a vital role in humour understanding. Contextual features that capture the surrounding text can be highly transferable for humour style classification, as different humour styles often rely on context for interpretation. This feature is especially useful for identifying affiliative humour style, where understanding the context of social interactions is key to recognising  humour and camaraderie, and also for identifying aggressive humour styles that rely on the context of criticism or ridicule. Hence, recognising the contextual cues, like a confrontational tone or sharp comments, is essential to discerning the aggressive humour style. 
        \item \textbf{Incongruity:} Incongruity, which represents the element of surprise or unexpectedness in humour, is a fundamental aspect of many humour styles. Models designed for sarcasm detection often consider incongruity. This makes the incongruity feature valuable for recognising aggressive humour styles, as this humour style is greatly based on sarcasm or mockery.
        \item \textbf{Ambiguity:}Ambiguity comprehension is crucial in humour recognition, especially for humour styles that involve wordplay or double meanings. This feature is relevant for humour styles like self-deprecating humour, where ambiguity or self-criticism can be used to create comedic effect.
        \item \textbf{Phonetic Style (Alliteration):} 
        Humour entails using phonetic features to produce comic effects, such as sounds, rhythms, wordplay, and speech patterns. This feature is applicable to self-enhancing humour, which may involve playful phonetic elements for self-affirmation.
        \item \textbf{Part-of-Speech:} Understanding the grammatical structure of text is important for many humour styles. Part-of-speech features can help identify linguistic patterns associated with different humour styles. This feature is useful for identifying self-deprecating or self-enhancing humour, as it often involves specific linguistic patterns and self-referential language like ``me", ``myself", ``I", ``we", and ``us".
        \item \textbf{N-gram: }N-gram features capture short sequences of words, which can be relevant for identifying recurring linguistic patterns characteristic of certain humour styles. This feature is valuable for recognising positive sentiment humour styles like self-enhancing and affiliative styles that may involve recurring phrases or inside jokes among a group of friends.
        \item \textbf{Semantic Similarity: }Semantic similarity measures can help determine the relatedness of words or phrases, which is essential for humour styles involving semantic nuances and word associations. This feature is transferable to the humour styles classification,  as it is important for understanding the underlying sentiment in self-enhancing humour and identifying affiliative humour styles that rely on shared meanings.
        \item \textbf{Sentiment-based and Polarity: } The four humour styles are greatly reliant on sentiment and polarity, given that each humour style can be classified as either having a positive or negative sentiment. For instance, this feature is relevant for negative humour styles (self-deprecating and aggressive), where sentiment reversal and criticism are part of the humour.
        \item \textbf{Acoustic Features:} The acoustic features that may be transferable for the effective identification of different humour styles are pitch and tone, timing and rhythm, and voice modulation. The acoustic features related to pitch and tone can convey emotional nuances in speech. These features can be relevant for identifying self-enhancing humour, which often involves cheerful and upbeat delivery to enhance one's mood. The timing and rhythm of speech or delivery can influence the perception of humour. These features can be applied to humour styles like self-deprecating and affiliative humour, where timing and rhythm play a role in comedic timing or creating a friendly atmosphere. Voice modulation, including changes in volume and intensity, can be relevant for aggressive humour styles as it may reflect sarcasm or mockery.
        \item \textbf{Visual Features:} Visual features have been shown to be effective in emotion recognition. Given that the four humour styles are associated with emotions, this makes visual cues valuable for the identification of the various humour styles. The visual features that are transferable from the reviewed papers are facial expressions, gestures, body language, and visual context. Visual features related to facial expressions, such as smiles and laughter, can be indicative of self-enhancing and affiliative humour styles, where humour is often expressed through positive facial cues. Visual cues like sad or sarcastic facial expressions can be indicative of self-deprecating and aggressive humour styles, where humour is often expressed through negative facial cues. Gestures and body language can provide context for humour styles, especially in self-deprecating humour, where individuals may use physical gestures to complement their humour. The visual context of a scene or image can influence the perception of humour. Visual context can be relevant for identifying negative humour styles (self-deprecating and aggressive) in visual content, where the context may contain elements of mockery or sarcasm.
    \end{enumerate}
    
\subsection{Transferable Computational Models} 
Based on the analysis of the included research papers, it is evident that a broad spectrum of existing models, ranging from traditional machine learning models to specialised task-specific models, can be suitably adapted for the task of humour style classification. This adaptability is supported by several compelling reasons:
\begin{enumerate}
        \item \textbf{Traditional ML Models:} Traditional ML models like SVM, NB, RF, DT, LR, and KNN are highly adaptable to different feature sets. They can utilise a wide range of linguistic and contextual features, which are also relevant for humour style classification. Furthermore, these models are often more interpretable, making it easier to analyse which features contribute to the classification decision. This interpretability can help in understanding how different features relate to humour styles.
        \item \textbf{Neural Network Models:} Neural network models, particularly recurrent models like LSTM, RNN, and GRU, are adept at capturing sequential and contextual information, which is vital for humour style understanding. They can capture dependencies between words and phrases that contribute to the identification of humour styles.  Similarly, neural networks are highly flexible and can be adapted to different input data and tasks. This adaptability makes them suitable for the diverse characteristics of humour styles.
        \item \textbf{Transformer Models: }Transformer models have been identified as excelling at  capturing contextual information and semantic relationships within text. They can comprehend the subtleties and nuances often associated with humour styles. In addition, pre-trained transformer models, such as BERT and RoBERTa, have learned rich linguistic representations from vast amounts of text data. Fine-tuning these models for humour style classification can leverage this pre-trained knowledge.
        \item \textbf{Specialised Models: }Specialised models are designed with humour, sarcasm, or sentiment analysis tasks in mind. They often incorporate domain-specific knowledge and architectural components that can be directly relevant to humour style classification. Moreover, some specialised models, like MSAM or MISRMSA, are designed for multimodal sentiment analysis. Given that humour styles can be expressed through text, audio, and visuals, these models can handle the multimodal nature of humour style recognition.
    \end{enumerate}
In summary, these existing models are transferable for humour style classification because they possess essential qualities such as adaptability, contextual understanding, and task-specific expertise. By fine-tuning these models with relevant datasets and features specific to humour styles, researchers and practitioners can leverage their strengths to recognise and classify humour styles effectively. However, it is crucial to carefully design experiments, pre-process data, and select appropriate features to maximise their transferability.

\section{Research Gaps and Opportunities}
Within the domain of computational-based approaches for humour style classification and related tasks, this SLR has revealed a spectrum of research gaps and promising opportunities for future investigations. Below, we present a consolidated overview of these identified research gaps and potential directions for future research.
\begin{enumerate}
    \item \textbf{Diverse and Balanced Datasets:} Existing research heavily relies on social media, in particular X (formerly Twitter), and Reddit  and English-language datasets, which may not capture the full spectrum of humour across languages and cultures. Future research should focus on constructing diverse and balanced datasets that encompass various humour types and languages, reducing dataset bias. Currently, there is a scarcity of comprehensive humour style datasets that cover all four primary humour styles: affiliative, self-deprecating, aggressive, and self-enhancing. The absence of such datasets hinders the development and evaluation of computational models for humour style classification.
    \item \textbf{Contextual Analysis of Humour Styles:} Existing studies have not explored humour style classification extensively in the context of mental health and relationships. Future research should investigate how humour styles, especially affiliative and self-enhancing styles, can impact individuals' well-being, self-esteem, intimacy, and mood in relationships and mental health.
    \item \textbf{Subjectivity and Labeling Bias:} Humour is inherently subjective and context-dependent, posing challenges to accurately labelling datasets. Future research should address the subjectivity of labelling humour styles and their related tasks and develop techniques to mitigate labelling bias, ensuring the quality of annotated datasets. 
    \item \textbf{Exploration of Understudied Humour Styles:}  While self-deprecating humour has received significant attention, there is a lack of research into the less explored humour styles, such as aggressive, affiliative, and self-enhancing. Investigating these understudied styles is vital for a comprehensive understanding of humour.
    \item \textbf{Multiclass Classification Models:}	 Current research predominantly relies on binary classification for humour. Future investigations should focus on the development of multiclass classification models that differentiate between affiliative, self-deprecating, aggressive, and self-enhancing humour styles.
    \item \textbf{Multimodal Data Integration:}  Humour often involves multiple modalities, including text, audio, video, and images. Future investigations should explore effective strategies for integrating these modalities seamlessly to achieve a holistic understanding of humour.
    \item \textbf{Specific Feature Identification: } To improve classification accuracy, it is essential to identify features specific to different forms of humour, such as sarcasm, irony, and wordplay. Tailoring features to each form of humour can enhance the precision of classification models. 
    \item \textbf{Context Detection:} Developing models capable of identifying appropriate contexts for humour and sarcasm usage is crucial for natural human-machine communication. Understanding when and where humour is suitable enhances the user experience. 
    \item \textbf{Cross-Cultural and Multilingual Analysis:} Investigating the cross-cultural and multilingual aspects of humour styles and its related-task can lead to better translation and interpretation of humour across languages and cultures. This research direction involves creating language-specific datasets and exploring how humour styles vary globally.
    \item \textbf{Enhanced Feature Engineering:}	Future research should explore advanced feature engineering techniques that consider not only linguistic features but also common sense, background knowledge, and emerging text modalities like emoji to improve humour style classification. 
    \item \textbf{Domain-Aware Datasets:} Researchers can work on creating domain-specific humour datasets to investigate the adaptation of humour styles in different contexts, such as humour in legal documents, medical reports, or technical literature.
    \item \textbf{Humour Style-Based Interventions:} Future research can explore the potential of using humour style-based interventions in mental health and relationship counselling. Developing AI systems that adapt humour styles to individuals' needs and preferences can be beneficial.
\end{enumerate}
In conclusion, the field of humour style classification and related tasks offers numerous research gaps and exciting opportunities for future investigations. Developing comprehensive datasets, exploring understudied humour styles, mitigating labelling bias, and improving contextual understanding are crucial steps towards a more comprehensive and nuanced analysis of humour styles and its related-tasks and their implications in various domains, including mental health and relationships.

\section{Conclusion}
This SLR has illuminated the dynamic landscape of computational-based approaches to humour style classification and related tasks. It encapsulated not only the richness of features and model advancements but also the enduring challenges of subjectivity, bias, and multimodal integration. Crucially, it underscored research gaps calling for comprehensive humour style datasets, exploration of contextual dimensions, multiclass classification, seamless multimodal integration, and cross-cultural analysis. The SLR serves as a clarion call to further explore this evolving field, advancing nuanced computational models and applications with profound implications. Moreover, it identified features and computational models for related tasks (binary humour and sarcasm detection) that are transferable for humour style classification tasks. Features include incongruity, sentiment and polarity, ambiguity, acoustic, visual, contextual information, and more, while computational models encompass traditional ML models, neural network models, transformer-based models, and specialised models. Lastly, it also identified existing humour and sarcasm datasets, providing easy access to resources for future research endeavours.

\printbibliography 

\appendix
\section{Appendices}
\subsection{Review of Primary Papers}

\begin{longtable}
{p{0.05\linewidth}| p{0.1\linewidth}| p{0.158\linewidth}| p{0.194\linewidth}| p{0.226\linewidth}| p{0.15\linewidth}}

\caption{Reviewed Papers \label{appendix:reviewed-papers}}
\\ \hline

\textbf{Paper} & \textbf{Task} 
& \textbf{Dataset} & \textbf{Feature} & \textbf{Computational Model} & \textbf{Performance Metrics} \\ \hline
\endfirsthead

\caption{Reviewed Papers (Continued)} \\ \hline
\textbf{Paper} & \textbf{Task} 
& \textbf{Dataset} & \textbf{Feature} & \textbf{Computational Model} & \textbf{Performance Metrics} \\ \hline
\endhead

    \cite{Patro2021MultimodalSitcoms} & Humour & MHD & Deep, visual & BERT, LSTM, Multi-modal Self Attention &  F1-Score, Accuracy, ROC\\ \hline
    \cite{Hasan2021HumorHumor} & Humour & UR-FUNNY, MUStARD  & Sentiment, Ambiguity, Acoustic, Visual & ALBERT, BERT, XLNet, Humour Knowledge Enriched Transformer(HKT) & Accuracy\\ \hline
    \cite{Liu2022FunnyNet:Videos} & Humour & MHD, MUStARD, UR-FUNNY, Friends &Deep, Acoustic, Visual &  FunnyNet (Cross-Attention Fusion) & F1-Score, Accuracy, Precision, Recall\\ \hline
    \cite{Castro2019TowardsPaper} & Sarcasm & MUStARD & Deep, Acoustic, Visual&  Random Majority, SVM & F1-Score, Precision, Recall\\ \hline
    \cite{Chauhan2022AnDetection} & Sarcasm & SEEmoji MUStARD & Word embedding, Acoustic, Visual, emoji Embedding & RNN (Gated Multimodal Unit) & F1-Score, Precision, Recall\\ \hline
    \cite{Christ2022MultimodalResults} & Humour Styles & Passau-SFCH & Deep, Acoustic, Visual & Gated Recurrent Units (GRUs), SVM & AUC\\ \hline
    \cite{Xu2022HybridDetection} & Humour & Passau-SFCH & Deep (BERT), Acoustic (DeepSpectrum), Visual (VGGface2) & Transformer-BiLSTM, BiLSTM & AUC \\ \hline
    \cite{Kathan2022AFairness} & Humour & Passau-SFCH & Acoustic (eGeMAPS, DeepSpectrum), Visual (VGGface2, Facial Action Units) & GRUs & AUC \\ \hline
    \cite{Christ2022TheStress} & Humour & Passau-SFCH & Deep (BERT), Acoustic (DeepSpectrum), Visual (VGGface2) & LSTM & AUC\\ \hline
    \cite{KamrulHasan2019UR-FUNNY:Humor} & Humour & UR-FUNNY & Word embedding (Glove), Acoustic (COVAREP), Visual (OpenFace) & Contextual Extension of memory fusion Network (C-MFN) & Accuracy \\ \hline
    \cite{Choube2020PunchlineFusion} & Humour & UR-FUNNY & Word embedding (Glove), Acoustic (COVAREP), Visual (OpenFace) & Context-Aware Hierarchical Network (CAHN)  & Precision, F1-score, Recall, Accuracy\\ \hline
    \cite{Sharma2020SemEval-2020Metaphor} & Humour, Sarcasm, offensive, motivation & Memotion & Word-embedding (Glove), Deep (CNN) & CNN-LSTM  & F1-Score \\ \hline

    \cite{Li2022Memeplate:Templates} & Humour & Memeplate & Deep  & BERT, ResNet-50, BEit,Faster-RCNN, XCit, WenLan & Accuracy, F1-score \\ \hline
    \cite{Chauhan2020All-in-One:Memes} & Humour, Sarcasm, offensive, motivation  & Memotion & Deep (BERT, ResNet) & Attention deep model &  F1-Score, Precision, Recall\\ \hline
    \cite{Bedi2021Multi-modalConversations} & Humour, sarcasm  & MaSaC & Word Embedding (FastText), Acoustic (Librosa) & LSTM  & F1-Score, Precision, Recall, Accuracy\\ \hline
    \cite{Joshi2016HarnessingFriends} & sarcasm  & Friends & N-gram, Punctuations, Sentiment & Naive Bayes, SVM & Precision, Recall, F1-score \\ \hline    
    \cite{Chauhan2021M2H2:Conversations} & Humour & M2H2 & Word embedding (fastText), Visual (ResNet), acoustic (openSMILE) & Dialogue-RNN, LSTM, Modality-Invariant and-Specific Representations for Multimodal Sentiment Analysis & Precision, Recall, F1-score\\ \hline
    \cite{Das2018SarcasmCNN} & Sarcasm  & Yahoo Flickr Sarcasm & Deep  & CNN & Accuracy\\ \hline
    \cite{Kamal2020Self-deprecatingApproach} & Humour style & E-tweets & Self-deprecating, Exaggeration, Word Embedding, ambiguity, polarity, subjectivity, phonetic style & RF & Precision, Recall, F1-score \\ \hline  
    \cite{Abulaish2019Self-DeprecatingApproach} & Sarcasm (Humour style) & E-tweets \cite{Ptacek2014SarcasmTwitter} &Punctuation Features, POS & DT, NB, Bagging & Precision, Recall, F1-score \\ \hline
    \cite{Kamal2019AnData} & Sarcasm (Humour style) & E-tweets \cite{Ptacek2014SarcasmTwitter} & N-gram Features, POS & LSTM & Precision, Recall, F1-score \\ \hline
    \cite{Annamoradnejad2020ColBERT:Humor} & Humour & ColBert & Incongruity, Deep  &  DT, SVM, XGBoost, XLNet, BERT & Accuracy, precision, recall, F1-score\\ \hline
    \cite{Weller2020TheCollection} & Humour & rJokes & Deep feature & BERT, roBERTa, XLNet & RMSE, Pearson, spearman \\ \hline
    \cite{Huang2022SICKNet:Knowledge} & Humour & rJokes, English Plaintext Jokes &Incongruity, Contextual information, Deep & BERT, RoBERTa, Semantic Incongruity and Commonsense Knowledge (SICKNet)  & Accuracy, F1-score \\ \hline
    \cite{Ptacek2014SarcasmTwitter} & Sarcasm & Tweets & N-gram, punctuation, emoticons, POS  & SVM & F1-score, conﬁdence interval \\ \hline    
    \cite{Ren2020SarcasmNetwork} & Sarcasm & Tweets \cite{Ptacek2014SarcasmTwitter}, IAC \cite{Oraby2017CreatingDialogue} & Sentiment, Deep, word embedding & Multi-level memory network based on sentiment semantics & F1-score, precision, recall \\ \hline
  
    \cite{Oliveira2020CorporaPortuguese} & Humour & Portuguese jokes & TF-IDF, N-gram, polarity, phonetic style (alliteration), ambiguity, Incongruity, Named Entities  & SVM, RF &F1-score, recall, precision  \\ \hline
    \cite{Abu-Farha2020FromDataset} & Sarcasm, Sentiment & Arsarcasm & Deep, sentiment, subjectivity & BiLSTM & F1-score, recall, precision \\ \hline
    \cite{Oprea2019ISarcasm:Sarcasm} & Sarcasm & iSarcasm & Deep & LSTM, CNN, Multi-dimension Intra-Attention Network (MIARN) & F1-score, precision, recall \\ \hline
    \cite{Diao2020ADetection} & Sarcasm & IACv2 & Word embedding, OOV words&   Muti-dimension attention-based BiLSTM  & Precision, Recall, F1-measure\\ \hline
     \cite{Kumar2020SarcasmLSTM} & Sarcasm & SARC & word embedding, sentiment, punctuation  & SVM, Multi-Head self-Attention based BiLSTM & F1-score, precision, recall\\ \hline     
    \cite{Ahuja2022Transformer-BasedIrony} & Sarcasm & SARC \cite{Khodak2018ASarcasm}, Tweet-Sarcasm \cite{Riloff2013SarcasmSituation} & Word embedding, TF-IDF  & CNN  & Accuracy, f1-score, precision, recall, AUC \\ \hline
     \cite{Jariwala2020OptimalHeadlines} & Sarcasm & News headlines & N-gram, POS, polarity, sentiment, punctuation,  & SVM & Accuracy, f1-score, precision, recall \\ \hline  
    \cite{Misra2023SarcasmDataset} & Sarcasm & News headlines & Word embedding, N-gram  & Attention-based Hybrid Neural Network & Accuracy  \\ \hline
    \cite{Kamal2022CAT-BiGRU:Detection} & Sarcasm & Twitter-280 & Self-deprecating pattern, word embedding, n-gram  & Convolution and Attention with BiGRU & Accuracy, f1-score, precision, recall\\ \hline
    \cite{Blinov2019LargeRecognition} & Humour & Stierlitz and FUN & TF-IDF, Deep features & SVM, Universal Language Model Fine-tuning (ULMFiT) & F1-score, recall \\ \hline 
    \cite{Fan2020HumorNetwork} & Humour & 1600 One-liner \cite{Mihalcea2005MakingRecognition}, Pun of the Day \cite{Yang2015HumorExtraction} & Word embedding, ambiguity & Internal and external attention neural network  & Accuracy, precision, f1-score, recall  \\ \hline
    \cite{Fan2020PhoneticsRecognition} & Humour & 1600 One-liner \cite{Mihalcea2005MakingRecognition}, Pun of the Day \cite{Yang2015HumorExtraction} &Ambiguity, Phonetic style, Deep, Word embedding & Phonetics and Ambiguity Comprehension Gated Attention network (PACGA)  & Precision, Recall, and F1-score  \\ \hline
    \cite{Zhang2017InvestigationsRecognition} & Humour & 1600 One-liner & Contextual knowledge, polarity, subjectivity, phonetic, incongruity, ambiguity & Attention-based BiGRU & Precision, recall, f1-measure \\ \hline
    \cite{Ren2021ABML:Detection} & Humour & 1600 One-liner & POS, word embedding  & Multi-Head Attention (MHA) &Precision, recall, F1-score \\ \hline
    \cite{Yang2015HumorExtraction} & Humour & Pun of the Day & Incongruity, Ambiguity, Phonetic Style, Polarity, subjectivity, BOW, Word embedding & KNN &Accuracy, f1-score, precision, recall  \\ \hline
    \cite{Ziser2020HumorSystems} & Humour & Product Question Answering &Deep, Incongruity, Subjectivity, N-gram  & LR, NB, LSTM, CNN & Accuracy, Recall \\ \hline
    \cite{Meaney2021SemEval-2021Offense} & Humour & SemEval-2021 Task 7 & BOW, TF-IDF, Deep & NB, SVR, BERT & Accuracy, F1-score \\ \hline
    \cite{Xie2021UncertaintyRecognition} & Humour & SemEval-2021 Task 7 & Word embedding, Semantic similarity, Phonetic style, ambiguity & SVM, GPT-2 & F1-score, precision, recall, accuracy\\ \hline
    \cite{Morales2017IdentifyingSources} & Humour & Yelp Dataset & Incongruity, Ambiguity, Phonetic Style (Alliteration) & NB, AdaBoost, Perceptron & Accuracy \\ \hline
    \cite{Hossain2019PresidentHeadlines} & Humour & Humicro-edit Dataset & N-gram, word embedding (GloVe)  & RF, LSTM & Accuracy\\ \hline    
    \cite{Cao2021Self-AttentionAssessment} & Humour & Humicro-edit Dataset  & Word embedding (GloVe), Incongruity, Deep (BERT) & Multi-Head Attention (MHA),RoBERTa, BERT & RMSE\\ \hline
    \cite{Alexandru2021TracingHeadlines} & Humour & Humicro-edit Dataset & TF-IDF & NB, SVM, NN & Accuracy \\ \hline
    \cite{Swami2018ADetection} & Sarcasm & English- Hindi code-mixed & N-gram, Emoticons  & SVM, RF & F1-score \\ \hline
    \cite{Pandey2023BERT-LSTMPost} & Sarcasm & English- Hindi code-mixed & TF-IDF, word embedding (GloVe), Deep (BERT)  & NB, SVM, RF, DT, KNN, LR, DNN, CNN, LSTM, BERT,   BERT-LSTM &  F1-score, precision, recall \\ \hline   
    \cite{Tang2022TheHumor} & Humour & Naughty-former & Deep Feature  & BERT, RoBERTa, DeBERTa  & Accuracy, precision, recall, F1-score \\ \hline
    \cite{Oraby2017CreatingDialogue} & Sarcasm & IAC & N-gram, Word embedding (Word2Vec) & SVM &  Precision, Recall, F1-Score \\ \hline
    \cite{Ren2020SarcasmNetwork}& Sarcasm & IAC, Tweets\cite{Ptacek2014SarcasmTwitter} & Word Embedding (GloVe), Sentiment, Deep feature & MMNSS & Precision, Recall, F1-Score \\ \hline
    \cite{Ghosh2016FrackingNetwork} & Sarcasm & Sarcasm-Detection & BOW, POS, Sentiment, Deep & Recursive-SVM, CNN, LSTM, CNN+LSTM+
    DNN & Precision, Recall, F1-Score\\ \hline    
    \cite{Chia2021MachineDetection}& Sarcasm & Sarcasm-Detection & TF-IDF, word embedding & CNN & f1-score \\ \hline
    \cite{Potamias2020ADetection}& Sarcasm & Sarcasm-Detection, Tweet-Sarcasm & Deep feature & Combined RCNN and RoBERTa & Precision, recall, accuracy, AUC, f1-score \\ \hline 
    \cite{Eke2021Context-BasedModel}& Sarcasm & Sarcasm-Detection, Tweet-Sarcasm & Word embedding (GloVe), Deep (BERT), Sentiment,  & Bi-LSTM & Precision, recall, accuracy, AUC, f1-score \\ \hline 
    \cite{Riloff2013SarcasmSituation} & Sarcasm & Tweet-Sarcasm & Sentiment, N-gram & SVM & Precision, recall, f1-score\\ \hline 
    \cite{Ahuja2022Transformer-BasedIrony}& Sarcasm & Tweet-Sarcasm, SARC & Word embedding (GloVe), TF-IDF & NB, LR, SVM, RF, KNN,CNN, LSTM, GRU, LSTM-AM, GRU-AM, BERT, ELECTRA, XLNet, RoBERTa, XLM-RoBERTa, ULMFIT & Precision, recall, accuracy, AUC, f1-score\\ \hline

 \end{longtable}

\end{document}